\documentclass[sigconf]{acmart}

\usepackage[font={small,it}]{caption}
\usepackage{graphicx}
\usepackage[utf8]{inputenc} 
\usepackage[T1]{fontenc}    
\usepackage{hyperref}       
\usepackage{url}            
\usepackage{amsfonts}       
\usepackage{nicefrac}       
\usepackage{microtype}      
\usepackage{amsmath}
\usepackage{subcaption}
\usepackage{algorithm}
\usepackage{algpseudocode}
\usepackage{geometry}

\setcopyright{rightsretained}

\copyrightyear{2024}
\acmYear{2024}
\setcopyright{rightsretained}
\acmConference[ICVGIP 2024]{Indian Conference on Computer Vision Graphics and Image Processing}{December 13--15, 2024}{Bengaluru, India}
\acmBooktitle{Indian Conference on Computer Vision Graphics and Image Processing (ICVGIP 2024), December 13--15, 2024, Bengaluru, India}
\acmPrice{}
\acmDOI{10.1145/3702250.3702281}
\acmISBN{979-8-4007-1075-9/24/12}

\begin{document}
\title{Variational Distribution and Experience Replay for 3D Reconstruction in a Continual Learning Framework}
\titlenote{Produces the permission block, and
  copyright information}

\author{Sanchar Palit}
\affiliation{%
  \institution{Indian Institute of Technology Bombay}
  \city{Mumbai}
  \country{India}}
  \orcid{0009-0002-3007-9330}
\email{204070004@iitb.ac.in}

\author{Sandika Biswas}
\affiliation{%
  \institution{Monash University}
  \city{Melbourne}
  \country{Australia}}
\email{sandika.biswas@monash.edu}


\renewcommand{\shortauthors}{}

\begin{abstract}
Single-image 3D reconstruction is a research challenge focused on predicting 3D object shapes from single-view images, requiring all training data for all objects to be available from the start. In dynamic environments, it's impractical to gather data for all objects at once; data becomes available in phases with restrictions on past data access. Therefore, the model must reconstruct new objects while retaining the ability to reconstruct previous objects without accessing prior data.
Additionally, existing 3D reconstruction methods in continual learning fail to reproduce previous shapes accurately, as they are not designed to manage changing shape information in dynamic scenes. To this end, we propose a continual learning-based 3D reconstruction method. Our goal is to design a model that can accurately reconstruct previously seen classes even after training on new ones, ensuring faithful reconstruction of both current and previous objects. To achieve this, we propose using variational distribution from the latent space, which represent abstract shapes and effectively retain shape information within a simplified code structure that requires minimal memory. Additionally, saliency maps preserve object attributes, capturing both local minor shape details and the overall shape structure. We employ experience replay to leverage these saliency maps effectively. Both methods ensure that the shape is faithfully reconstructed, preserving all minor details from the previous dataset. This is vital due to resource constraints in storing extensive training data. Thorough experiments show competitive results compared to established methods, both quantitatively and qualitatively.




\end{abstract}

%
%
\ccsdesc{Computer vision problems~shape reconstruction}
\ccsdesc{Machine Learning settings~Continual learning settings}

\keywords{3D reconstruction, Continual Learning, Variational Inference, Image saliency}

\maketitle

\section{Introduction}
\label{sec:intro}  


Single-image 3D reconstruction~\cite{mescheder2019occupancy,choy20163d,maturana2015voxnet,xie2019pix2vox,wang2018pixel2mesh} aims to generate a 3D model from a single RGB image, representing an object or scene in mesh\cite{wang2018pixel2mesh}, point cloud\cite{mescheder2019occupancy}, or voxel\cite{xie2019pix2vox} form. 
Learning-based single-image 3D reconstruction models depend significantly on extensive datasets for training, which limits their applicability to a wide range of incoming classes.
Recent advances in learning-based methods show promise in this field, with some category-specific 3D reconstructions \cite{xie2019pix2vox, choy20163d} excelling on specific classes. 
Other approaches leverage 3D-CNNs and GANs \cite{wu2016learning} to synthesize 3D objects in the latent space. 
\begin{figure}
  \centering
   \includegraphics[width=1.0\columnwidth]{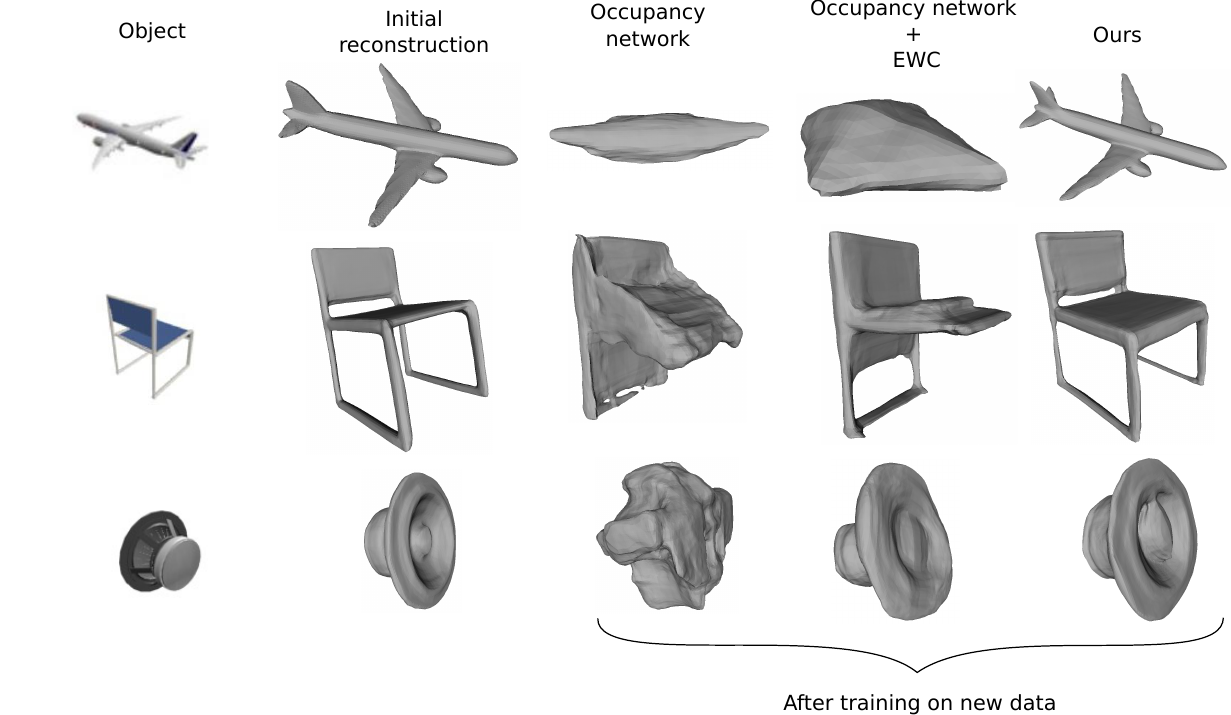}
   \caption{\footnotesize{This figure shows the catastrophic forgetting in learning-based 3D reconstruction models, such as Occupancy network\cite{mescheder2019occupancy}. Initially, the model is trained on classes such as airplane, chair, and speaker, successfully reconstructing these shapes. However, after training the model on a new object, such as a car, while the model is able to reconstruct the car, the reconstruction of the previously learned shapes becomes inaccurate, as demonstrated (with and without EWC\cite{kirkpatrick2017overcoming}, a continual learning based method). In contrast, Our method is able to faithfully reproduce the shape even after training the model on new objects thus mitigating catastrophic forgetting.}}
   \label{fig:initial}
\end{figure}
These methods require the availability of training data for all incoming objects at the outset, which is often impractical. To address this limitation, we propose a continual learning framework that trains the model using a stream of distinct incoming object classes, ensuring non-overlapping objects between sessions. Initially, we train the model with a subset of objects. Subsequently, we continue training the model on new sets of data at different instances, without access to the previous training data. Despite this, the model remains capable of faithfully reconstructing both previously seen objects and newly encountered ones, a feat that existing methods fail to achieve. 

Continual learning, extensively studied in the literature \cite{kirkpatrick2017overcoming, yoon2017lifelong, lopez2017gradient}, tackles the problem of catastrophic forgetting in standard computer vision models. By adopting continual learning, the model can update and refine its 3D representation over time as new data becomes available. This enables the model to retain knowledge of previously seen objects while effectively learning new objects in subsequent sessions. As a result, the model faithfully reproduces both previously encountered and current objects after being trained on the current dataset, incrementally expanding its knowledge without forgetting previously learned shapes.

In the existing literature, training a 3D reconstruction model on new classes can result in the loss of its ability to faithfully reconstruct previously learned objects. This highlights the necessity for a continual learning framework in this context, which remains unexplored. In the context of continual learning, existing 3D reconstruction models have some potential drawbacks. Occupancy networks \cite{mescheder2019occupancy}, designed for static scenes, will not be able to handle dynamic scenes with moving or changing objects well. As more data is added, the memory requirements of occupancy networks can become prohibitive, particularly for large-scale 3D shapes. Without regularization constraints, training an occupancy network on a new dataset may result in catastrophic forgetting, leading to inaccurate reconstructions of previously encountered shapes.
3D-R2N2 \cite{choy20163d}, trained on a fixed set of objects, is not easily adaptable to new objects. Adding new objects requires retraining the model from scratch, which is time-consuming and computationally expensive. Furthermore, directly integrating existing 3D reconstruction methods with current continual learning techniques often results in imperfect reconstructions. This incompatibility arises because continual learning methodologies are generally designed for handling classification tasks across numerous classes (e.g., occupancy net \cite{mescheder2019occupancy} and EWC \cite{kirkpatrick2017overcoming}), as shown in Figure~\ref{fig:initial}.
In contrast, 3D reconstruction demands highly supervised training for each individual class to accurately generate the shape of the object. To the best of our knowledge, this work is the first attempt to mitigate catastrophic forgetting in the context of single-image 3D reconstruction. Previous work has addressed the challenges of less data, few-shot categories\cite{xing2022few, wallace2019few, michalkiewicz2020few}, and unseen classes\cite{marton2010general}. 


Therefore, we propose a methodology to mitigate catastrophic forgetting while also providing a framework that can be judiciously extended to handle diverse class categories leveraging the prior knowledge from Variational priors. We encode each shape into latent distributions to effectively capture its configuration. These latent distributions, or variational priors, are highly effective and have been employed in our novel continual learning 3D reconstruction framework to retain previously learned knowledge. In practical situations, storing shapes in an abstract form or as a large training set becomes impractical due to resource constraints. However, Variational Priors offer an effective means to represent abstract shapes and address forgetting, as evidenced by Variational Continual Learning\cite{nguyen2017variational}. Additionally, we employ saliency map-based experience replay to store global and distinctive object attributes. This experience replay is highly effective in the context of continual learning. Saliency maps are used to preserve distinctive object characteristics and focus on the important parts of various objects. This combination of Variational Priors and saliency maps presents a promising approach for efficient and effective continual 3D reconstruction.

In summary, we propose three strategies to enhance the task:
\textit{Global Variational Distribution priors:} Introducing global priors to the variational distribution improves representation learning.
\textit{Global and Local Saliency of Images:} Leveraging global and local saliency information from images enhances shape reconstruction.
\textit{Composition of Variational priors:} Utilizing attention mechanisms effectively combines variational priors, leading to improved shape reconstruction.

\section{Related work}
\label{sec:related_work}


\subsection{Continual Learning:}

Continual Learning encompasses various categories, including Dynamic architectural methods, Memory-based methods, Regularization methods, Bayesian approaches, and Natural gradient descent methods. Memory-based methods \cite{lopez2017gradient, riemer2018learning, rebuffi2017icarl, shin2017continual} retain previously encountered data or create some representation of past data through generative modeling. Experience Replay-based methods \cite{riemer2018learning, chaudhry2019continual} jointly train the model using both previous and current samples. Regularization-based methods \cite{kirkpatrick2017overcoming, zenke2017continual, aljundi2019task} overcome catastrophic forgetting in fixed capacity models by penalizing significant changes in important model measures. Bayesian approaches \cite{nguyen2017variational, hernandez2015probabilistic} utilize Bayesian methods to produce posterior distributions of network weights, incorporating uncertainty measures for more accurate parameter estimates. Although these techniques are primarily designed for classification tasks, adapting them to 3D reconstruction methods would require additional annotation and overhead, introducing further challenges to the task. 

\subsection{Single-image 3D reconstruction:} 

Traditional single-image 3D reconstruction methodologies necessitate substantial data for model training. Structure from Motion (SFM) \cite{schonberger2016structure} and Simultaneous Localization and Mapping (SLAM) \cite{cadena2016past} offer viable strategies for data annotation. Present learning-based 3D reconstruction approaches are categorized into voxel-based, mesh-based, or point-based methods, according to their resulting output representations. Voxel-based techniques \cite{maturana2015voxnet, wu20153d, song2016deep} utilize either generative or discriminative frameworks, employing convolutional neural networks that operate on voxel grids. However, these methods have memory limitations, are confined to specific voxel grid sizes \cite{wu2017marrnet}, and would yield unsatisfactory outcomes when integrated into a continual learning context. Furthermore, generative models based on GANs or VAEs \cite{jimenez2016unsupervised, wu2016learning} are susceptible to mode collapse and divergent outcomes when employed with limited data. Our approach avoids the use of generative modeling, thereby minimizing additional computational overhead, and exhibits significantly reduced storage requirements compared to conventional memory-based techniques.

\subsection{3D reconstruction under less supervision:} 
Various approaches have endeavored to address the challenges of extensive data requirements and prolonged training times in 3D reconstruction. These attempts involve investigating semi-supervised settings \cite{xing2022semi, yang2018learning, bozic2020deepdeform} and few-shot scenarios \cite{michalkiewicz2020few, xing2022few}. Some studies \cite{zhang2018learning, shin2018pixels, wang2020gsir} have estimated depth and normals for extrapolation to unseen classes. Additionally, zero-shot single-image 3D reconstruction \cite{liu2023zero} aims to explore model generalization in the context of unobserved classes. Single-image 3D reconstruction demands robust shape priors for model training. However, incremental training with new data can result in the model forgetting prior knowledge. In our work, we not only propose a method to alleviate this forgetting but also devise an approach enabling the model to utilize newly acquired knowledge to enhance performance on previously learned shapes, termed as Backward Knowledge Transfer (BWT).



\subsection{Continual Learning under Constraints:}

While most recent research in Continual Learning has primarily focused on classification tasks, a growing body of work is delving into its application in other domains. These include tasks such as image segmentation \cite{michieli2019incremental, cermelli2020modeling, maracani2021recall}, 3D object detection \cite{yun2021defense, shmelkov2017incremental}, and generative tasks \cite{xiang2019incremental, lesort2019generative}. Furthermore, the exploration of Continual Learning under more challenging constraints, such as few-shot learning \cite{palit2022prototypical, tao2020few, zhang2021few} and semi-supervised scenarios \cite{zhang2006new}, has gained attention. There have been attempts to integrate Continual Learning with reinforcement learning \cite{abel2018state, xu2018reinforced}, showcasing synergies between the two paradigms. While there are certain overlaps with our work, such as \cite{yan2021continual} which focuses on scene reconstruction using signed distance fields for 3D shape representation, our approach introduces additional complexities. We address distinct object distributions, account for potential distribution shifts, and tackle long-tailed data scenarios. This sets our work apart and presents unique challenges in the realm of Continual Learning.

\section{Background:}
\label{sec:background}
\textbf{Variational Continual Learning:}
\label{subsec:vcl}
Variational Continual Learning amalgamates Variational Inference, which incorporates a Bayesian model represented as $\mathbb{P}(\mathcal{Y}/\mathcal{W}, \mathcal{X})$, utilizing network weight $\mathcal{W}$ to accommodate an input-output pair $\mathcal{D} = (\mathcal{X}, \mathcal{Y})$. 
The underlying presumption is that the weights are sampled from a prior distribution $\mathcal{W} \sim \mathbb{P}(\mathcal{W};\theta)$.

During the training procedure, the computation of the precise posterior using Bayes' rule becomes computationally intractable. To address this challenge, Variational Inference approximates the posterior by optimizing the subsequent objective,
\setlength{\belowdisplayskip}{0pt} 
\setlength{\abovedisplayskip}{0pt}  
$$
\mathcal{Q}(\mathcal{W}|\theta) = \arg \min_{\theta} \mathrm{KL}(\mathcal{Q}(\mathcal{W}|\theta)||\mathbb{P}(\mathcal{W}|\mathcal{D})) 
$$
or equivalently by minimizing the following ELBO loss function,
$$
{\mathcal{L}_{\mathrm{ELBO}}(\theta, \mathcal{Q})} = \underset{\textsc{Log-Evidence}}{\underline{-\mathbb{E}_{\mathcal{Q}(\mathcal{W})}[\log \mathbb{P}(\mathcal{D})]}} + \underset{\mathrm{KL-Divergence}}{\underline{\mathrm{KL}(\mathcal{Q}(\mathcal{W}|\theta)||\mathbb{P}(\mathcal{W}|\mathcal{D}))}}\label{eq:elbo}
$$
The log-evidence loss can be mathematically formulated as the binary cross-entropy loss, which quantifies the level of concordance between the data and the distribution of model weights. In the Continual Learning context, the loss function during an incremental session $t$ can be represented as follows, where $\mathcal{Q}(\mathcal{W}|\theta)$ is a Gaussian distribution with parameters $\theta = \{\mu, \sigma\}$:
$$
\mathbb{E}_{\mathcal{Q}(\mathcal{W}_t|\theta_{t})} [- \log \mathbb{P}(\mathcal{D}_t |\mathcal{W}_t)] + \mathrm{KL}(\mathcal{Q}(\mathcal{W}_{t-1}|\theta_{t-1})|| \mathcal{ Q}(\mathcal{ W}_t|\theta_{t}))$$

\section{Method}
\label{sec:method}
\begin{figure*}[!htb]
  \centering
   \includegraphics[width=1.0\linewidth]{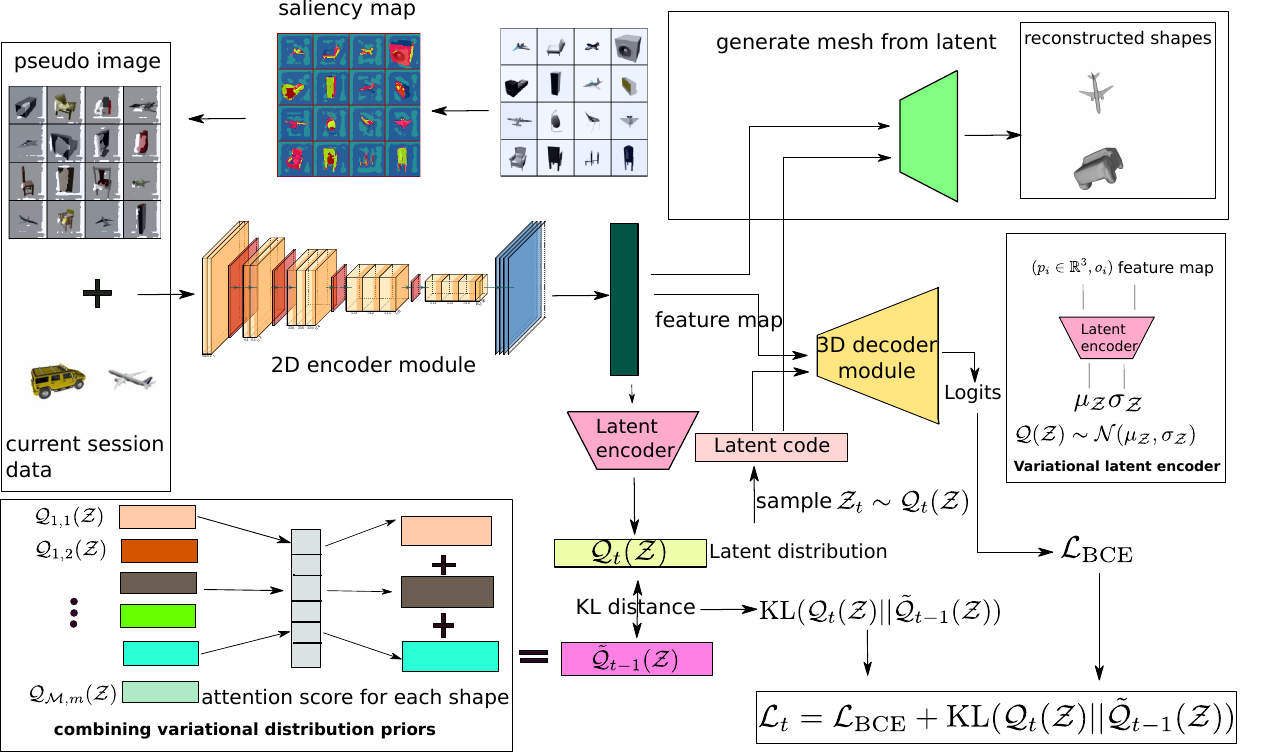}
   \caption{\footnotesize{During an incremental session, the training loss is updated as shown in the figure. The model is simultaneously trained using a combination of generated pseudo-images and the current dataset of the session. The variational latent encoder has acquired latent distributions corresponding to distinct shapes for each session. Denoting the latent distribution for the $j$-th class of the $i$-th session as $\mathcal{Q}_{i,j}(\mathcal{Z})$, and $\mathcal{Q}_{1,1}(\mathcal{Z}), \dots, \mathcal{Q}_{\mathcal{M}, m}(\mathcal{Z})$ as the latent distributions of shapes encountered in prior instances up to the $t-1$-th session, the mesh is generated from the latent code using Multiresolution IsoSurface Extraction.}}
   \label{fig:incremental_loss_update}
\end{figure*}

The problem in single-image 3D reconstruction entails generating the 3D shape representation, denoted as $\mathcal{S}$, based on a single image $\mathcal{I}$ captured from an arbitrary viewpoint.
In the Continual Learning 3D reconstruction framework, we deal with a sequential stream of incoming data denoted as $\mathcal{D} = \{\mathcal{D}_1, \dots \mathcal{D}_{\mathrm{T}}\}$ consisting of T-sessions where $t$-th session dataset is $\mathcal{D}_t = \{(\mathcal{I}_j^{t}, \mathcal{S}_j^{t})\}_{j=1}^{\mathcal{N}_t}$. Here, $\mathcal{I}_j^t$ denotes the input image of the $j$-th object, $\mathcal{S}_j^t$ corresponds to the 3D shape prior of the $j$-th object, and $\mathcal{N}_t$ represents the number of objects in session $t$. The established framework integrates a 2D image feature extractor network $\mathcal{E}:\mathcal{I} \rightarrow \mathrm{e}$, designed to extract pertinent image features, alongside a model termed the Latent encoder network $\mathcal{G}: \mathrm{e} \rightarrow \mathcal{Q}$. This network is responsible for learning latent distributions(or latent code) pertaining to individual shapes. The Latent encoder network functions by taking both ground truth and the feature map as inputs and subsequently generates the learned latent code for the shape. This latent code is represented by means and standard deviation values of a Gaussian distribution $\mathcal{Q}(\mathcal{Z})$, illustrated in Fig. \ref{fig:incremental_loss_update}.
In the context of standard Continual Learning, at session $t$ the model evaluation is conducted on an aggregated test dataset comprising samples from both current and prior sessions $\mathcal{C} = \{\mathcal{C}_1, \mathcal{C}_2, \dots \mathcal{C}_{t}\}$. However, during the training phase, the model's exposure is limited to the training dataset of the present session exclusively. 
Furthermore, a 3D decoder module $ \phi: \{e, \hat{\mathcal{Z}}\} \rightarrow \mathcal{O}$ is incorporated into the architecture where $\hat{\mathcal{Z}}$ denotes samples of the distribution $\hat{\mathcal{Z}} \sim \mathcal{Q}(\mathcal{Z})$. This module operates by accepting the feature map generated by the feature extractor and samples of the variational latent code as inputs. The outcome of this process is the probability distribution $\mathcal{O}$ associated with the 3D point's output. As a result, the trainable components within our methodology encompass the 2D encoder, the 3D decoder, and the variational latent code encoder.
 At any incremental session $t$ The model optimizes parameter $\theta_t = \{\mathcal{W}_{\mathcal{E}}, \mathcal{W}_{\mathcal{G}}, \mathcal{W}_{\phi}\}$ corresponding to $f(\mathcal{D}_t;\theta_t): \mathcal{I} \rightarrow \mathcal{O}$ by minimizing the loss function $\mathcal{L}_{t}( \mathcal{D}_t; \theta_t)$  where $f \equiv \{\mathcal{E}, \mathcal{G}, \phi\}$ during learning. To generate the shape $\Tilde{\mathcal{S}}$ corresponding to the image $\mathcal{I}$, an additional mesh generating module, which remains non-trainable, is employed. This module utilizes the latent code and the feature map as inputs in its operation. The overall process is shown in Fig.~\ref{fig:incremental_loss_update}

\textbf{Shape Encoding:} 
Let us analyze the encoder-decoder framework designed specifically for shape representation, where each shape is encoded into a latent vector.
Throughout the training process, a 2D encoder-3D decoder architecture, combined with the latent code, is employed to produce logits for individual shapes: 
$\mathcal{O} = \phi(\mathrm{e}_{\mathcal{I}}, \hat{\mathcal{Z}}) = \phi(\mathcal{E}(\mathcal{I}), \hat{\mathcal{Z}})$.
This setup enables the calculation of the binary cross-entropy loss.
To train this model, we employ a binary cross-entropy loss applied to the voxel output confidence $\mathcal{O}_i$ for voxel $i$ and its corresponding ground truth in the output grid, which can be expressed as:
$$\mathcal{L}_\mathrm{BCE} = -
\frac{1}{\mathrm{N}} \sum_{i=1}^{\mathrm{N}}\mathcal{Y}_i\log(\mathcal{O}_i)+(1-\mathcal{Y}_i)\log(1-\mathcal{O}_i)$$
Here, $\mathrm{N}$ represents the total count of voxels, and $\mathcal{Y}_i$ corresponds to the target ground truth occupancy value for the voxel indexed as $i$. The objective of this training procedure is to enhance the performance of the encoder-decoder framework, ensuring precise reconstruction of shapes using the given images and occupancy data.

Lastly, as previously elucidated, the process of shape reconstruction involves generating the shape utilizing the feature map and the sampled values from the latent distribution. This relationship can be expressed as follows:
$$\Bar{\mathcal{S}}= \texttt{GML}(\mathrm{e}_{\mathcal{I}}, \hat{\mathcal{Z}}) = \texttt{GML}(\mathcal{E}(\mathcal{I}), \hat{\mathcal{Z}})$$

Here, \texttt{GML}() is employed to generate the mesh using a standard mesh generation algorithm from voxel.

\textbf{Combining variational distribution priors:}
Object feature maps, as introduced in \cite{michalkiewicz2020few, wallace2019few}, provide a means to utilize class-specific training shapes for the acquisition of a representative shape prior. Nevertheless, the acquired embedding priors lack the explicit utilization of commonalities across diverse classes, leading to possibly suboptimal and duplicated representations. Despite this, it overlooks the potential to capitalize on shared traits among shapes, an approach that could enhance resilience in scenarios with changing data categories. 

To address this limitation, we propose an innovative strategy named Combining Variational Distribution Priors. This approach aims to learn compositional representations between classes, as depicted in Figure \ref{fig:incremental_loss_update}. Variational distribution priors enable the utilization of class-specific training shapes for the acquisition of a comprehensive shape prior. This phenomenon occurs because Variational Distributions possess dual strengths: they are not only potent enough to accurately capture the distribution of latent embeddings, but also exhibit an inherent lack of bias. Employing optimization-based variational inference \cite{zhang2018advances} offers the opportunity to match these distributions to any given posterior. 

Our primary goal is to actively prompt the model to uncover shared attributes inherent to various shapes, attributes that can be effectively applied across different instances. To achieve this, we propose the decomposition of our class-specific shape representation into a collection of variational prior samples that exhibit inter-class commonalities. 
To elaborate, we establish a set of $\Sigma_{t=1}^{T} m \cdot \mathcal{N}_t$ variational priors. 
For each object encountered up to the current session, we store $m$ distinct variational prior distributions. Each individual prior, obtained using the latent encoder as previously described and stored as $\{\mathcal{Q}_{i}(\mathcal{Z})\}_{i=1}^{\mathcal{N}_t}$, consists of $m$ distinct prior samples, $\mathcal{Q}_{i}(\mathcal{Z}) = \{\mathcal{Q}_{i,1}(\mathcal{Z}), \dots \mathcal{Q}_{i,m}(\mathcal{Z})\}$, where each $\hat{\mathcal{Z}} \in \mathbb{R}^{\mathrm{D}}$ with $\hat{\mathcal{Z}} \sim \mathcal{Q}_{i,m}(\mathcal{Z})$. Conceptually, each prior sample can be interpreted as a representation of an abstract concept that can be shared among multiple classes.

To augment the learning process, we introduce an additional KL divergence loss, which enforces proximity between the latent codes of newly acquired shapes and the latent code that can be generated by optimally combining those of previously learned shapes. This proximity is optimized using an attention mechanism that skillfully combines the latent codes of analogous shapes, thereby ensuring their convergence. Notably, our approach confines low memory usage by exclusively storing the latent codes of learned shapes. For training, we employ the loss function outlined in Variational Continual Learning \ref{subsec:vcl}. 
In a specific manner, we utilize the binary cross-entropy loss to quantify the log evidence from the current dataset and the Variational distribution priors of shapes obtained from previous sessions, which enables the computation of the KL distance.
The training procedure involves the integration of a loss akin to ELBO loss. 
The data log-likelihood is represented by a cross-entropy loss, while the KL divergence loss is employed to assess the dissimilarity between the latent variables obtained and the latent codes of shapes learned in earlier sessions.
Consequently, the comprehensive loss is formulated as follows:
$$ \mathcal{L} = \mathcal{L}_\mathrm{BCE} + \mathrm{KL}[\mathcal{Q}_t(\mathcal{Z})||\Tilde{\mathcal{Q}}_{t-1}(\mathcal{Z})]$$
For every session $t$, the attention vector identifies the most suitable distribution from the set of prior distributions up to session $t-1$. The computation of the latent prior with attention is expressed as follows:
$$ \Tilde{\mathcal{Q}}_{t-1}(\mathcal{Z}) = \sum_{k=1}^{\mathcal{M}} \sum_{j=1}^{m} \mathcal{A}_{t-1}^{k,j} \mathcal{Q}_{k,j}(\mathcal{Z})$$
In this context, $\mathcal{A}_{t-1}^{k,j}$ represents the attention scalar applied to the $j$-th variational prior sample within class $k$, and $\mathcal{Q}_{k,j}(\mathcal{Z})$ indicates the $j$-th variational prior sample of class $k$. The value of $\mathcal{A}^{k,j}$ is learned in each session, thus endowing it with variational prior sample-specific attributes.
\setlength{\belowdisplayskip}{0pt} 
\setlength{\abovedisplayskip}{0pt}  
$$\hat{\mathcal{A}}_{t-1}^{i} = \arg \min_{\mathcal{A}^{i}} \sum_{j=1}^{n} \mathcal{G}(\mathcal{E}(\mathcal{I}_j),\mathcal{Q}(\mathcal{Z}^{(i)}) $$
This process confers distinct attributes to individual variational priors. Our objective is to ensure that each attention importance allocates a discrete attribute to each variational prior, thus enabling the model to choose the relevant variational prior during a particular session.

\textbf{Experience replay using image saliency:}
Continual Learning models have demonstrated notable performance gains by preserving past samples and reintroducing them using diverse sampling criteria \cite{shim2021online, riemer2018learning, aljundi2019gradient} during the ongoing session. These past instances are stored in a memory buffer to alleviate the issue of catastrophic forgetting. The effectiveness of memory retention improves with a larger buffer size. However, the buffer's size presents a limitation within the context of online learning. Empirical evidence supports the notion that experience replay enhances model performance and diminishes catastrophic forgetting. We present a technique to store images in a significantly reduced-size buffer to accomplish this objective.

 Explainable AI \cite{van2008visualizing, zhou2016learning} endeavors to achieve internal reasoning within neural network models by utilizing tools like saliency maps\cite{selvaraju2017grad}. In line with this concept, we adopt saliency maps to extract crucial global and local features from various objects, enabling us to perform our reconstruction process better. This imparts us with a measure of prominence that guides our attention toward particular regions within the image. The utilization of the saliency map pertaining to prior objects facilitates the extraction of both local and global shape details. The saliency map for an image can be generated through various methodologies, as indicated by the following equation:
$\mathcal{I}_{\mathrm{SAL}} = \mathcal{SM} (\mathcal{I}, \mathcal{E}) \label{eq:saliency}$.
The process of obtaining the saliency map has been explored through various methods, and a comprehensive discussion is presented in Section \ref{sec:ablation}.
The saliency maps of distinct objects from the preceding session are stored in a dedicated memory module labeled as $\mathcal{M}_{\mathrm{SAL}}$.
Throughout incremental training, these prior saliency maps are reintroduced, and the model undergoes joint training on both the images of the current session and those corresponding to past saliency maps.
This approach bears semblance to methods rooted in Experience Replay\cite{hayes2019memory, riemer2018learning}. The preservation and utilization of both local and global saliency of images are embraced. In the case of local saliency, images are cropped to extract pivotal regions or significant portions. Subsequently, these localized saliency maps are interpolated to reconstruct the complete image, thus enabling image regeneration through appropriate mapping based on the saliency maps.
The saliency maps are reintroduced using the following formula:
$\mathcal{I}_{\mathrm{SAL}} \xrightarrow{\mathrm{interpolate}}\mathcal{I}_{\mathrm{pseudo}}$. Subsequently, the augmented pseudo-images are introduced into the current training dataset, and the joint training process is conducted, as illustrated in Fig.~\ref{fig:incremental_loss_update}.
We establish a thresholding mechanism on the saliency values to distinguish the pixels with significant importance. Subsequently, distinct regions of significance are visually highlighted through varying color codes, thereby assigning diverse levels of importance to different areas. To generate pseudo images, we utilize the saliency maps from different layers as blending tools for the original images. The blending process involves combining these elements to construct pseudo images based on the pixel values. This is accomplished by identifying pixel values within non-zero regions and subsequently replacing them with the original values for existing regions, while interpolating the average of neighboring values in intermediate regions. The specific choice of replacement depends on the nature of the objects. We either use the mean value of the different map values or assign a distinct color to create visually distinguishable pseudo images.
\begin{figure*}[!h]
  \centering
    {\includegraphics[width=3.3cm,height=3.3cm]{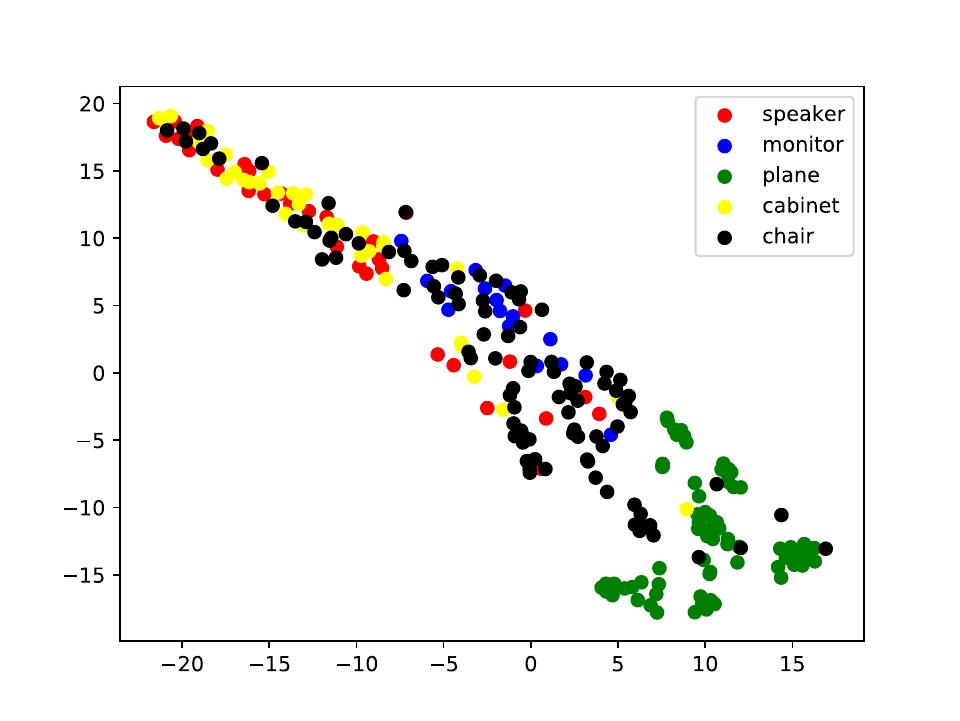}}
  \hfill
    {\includegraphics[width=3.3cm,height=3.3cm]{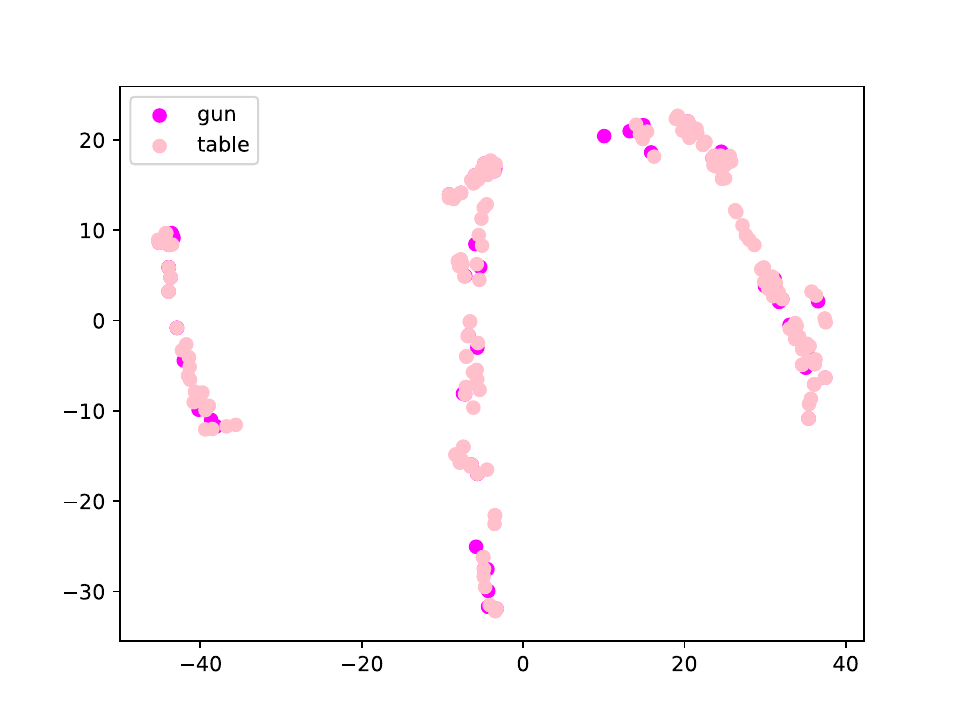}}
  \hfill
    {\includegraphics[width=3.3cm,height=3.3cm]{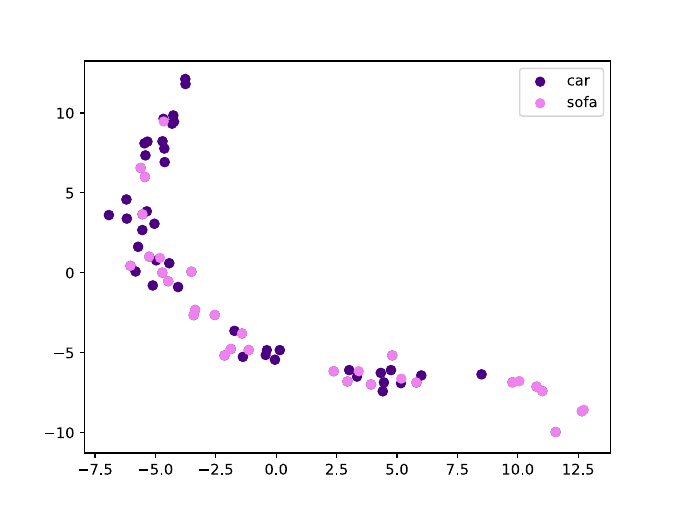}}
  \hfill
    {\includegraphics[width=3.3cm,height=3.3cm]{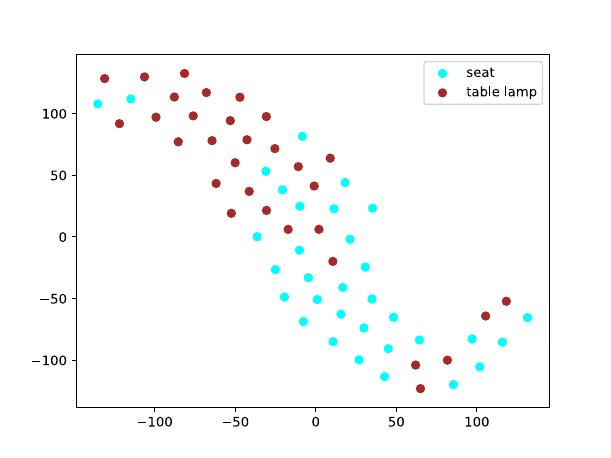}}
  \hfill
    {\includegraphics[width=3.3cm,height=3.3cm]{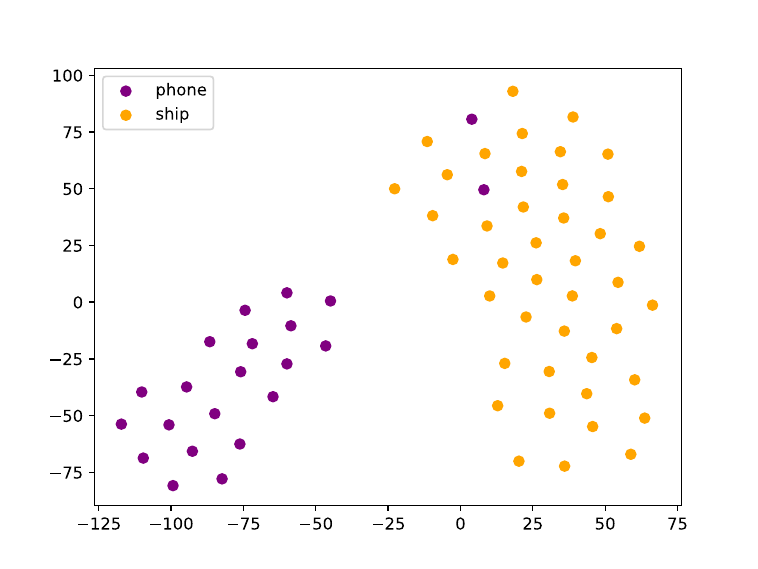}}
  \caption{Mean values of the latent variables for all objects from session 0 to 4 in the ShapeNet-13. Enlarge for a clearer view.}
  \label{fig:latent_mean}
\end{figure*}

\textbf{Inference:}
During inference, the \texttt{GML()} module, utilizing Multiresolution IsoSurface Extraction \cite{mescheder2019occupancy}, is employed for the mesh generation process. This module employs an octree structure to extract meshes from the outputs of the 2D-encoder and the variational latent encoder. 
The initial step involves discretizing the volumetric space to achieve meshes of higher resolution. Subsequently, grid points are classified as occupied if the corresponding encoder output exhibits a high value.


\section{Experiments:}
\label{sec:experiments}

\textbf{Dataset:} 
We use the ShapeNet-13 dataset \cite{choy20163d} for our experiments. It has 13 distinct categories and comprises a total of 43,783 3D models. The dataset is systematically divided into five sessions, with the 13 categories being incrementally assigned to these sessions. Additionally, for each session, the dataset is further segregated into training, validation, and test subsets. The voxel resolution utilized for ShapeNetCore.v2 is $32^3$. To compute the Intersection over Union (IOU) metric of the model, a suitable sampling strategy is applied from the original model to generate the ground truth. Additionally, the study includes findings from the KITTI dataset \cite{geiger2013vision} with annotations from \cite{zhang2015sensor} and \cite{ros2015vision}, encompassing a total of 10 categories.

\textbf{Implementation details}
For the ShapeNet dataset, the model is initially trained for 200 epochs, with subsequent incremental sessions being trained for 80 epochs each, with batch size 64. The KITTI dataset undergoes training for 80 epochs across all sessions. Moreover, a progressive reduction of the learning rate is incorporated at each session. In the initial phase of each incremental training session, the learning rate is set at $1e-3$, and subsequently reduced to $1/5$-th of its initial value after 25, 35, 45, and 55 epochs.

\textbf{Baselines:}
Our approach is systematically compared against various 3D-reconstruction baselines and extended versions of several continual learning methods \cite{lopez2017gradient, kirkpatrick2017overcoming, yoon2017lifelong} specifically designed for image classification tasks. For the purpose of this comparison, we employ a composite architecture comprising the 2D encoder-3D decoder, utilizing the architectural frameworks of Occupancy networks \cite{mescheder2019occupancy} and Signed Distance Function (SDF) networks \cite{thai20213d}. For the 2D image encoder, we employ the ResNet-18 architecture, which is further customized with additional parameter reductions and adjustments in the learning rate. Within the 2D image encoder, feature extraction is performed using ResNet blocks. To incorporate saliency from the input image, supplementary layers are incorporated. The decoder layer is comprised of 4 hidden layers with 512 units per layer. In the decoder portion, ResNet blocks coupled with conditional batch normalization are utilized.
 Additionally, we have incorporated conventional Continual Learning methods to address the challenge of forgetting in the networks. During incremental training, the parameters of both the feature extractor network and the latent code encoder module are updated using these methods. For a comprehensive evaluation, we extended the Occupancy Network with memory-based techniques such as Gradient Episodic Memory (GEM) \cite{lopez2017gradient}, regularization-based technique Elastic Weight Consolidation (EWC) \cite{kirkpatrick2017overcoming}, and dynamic architecture approach Dynamic Expansion Network (DEN) \cite{yoon2017lifelong}. To establish an upper limit of performance, we applied the Occupancy Network on all objects collectively.
\begin{figure*}
    \begin{minipage}{0.55\textwidth}%
        \includegraphics[width=1.35\textwidth, height=1.35\textwidth]{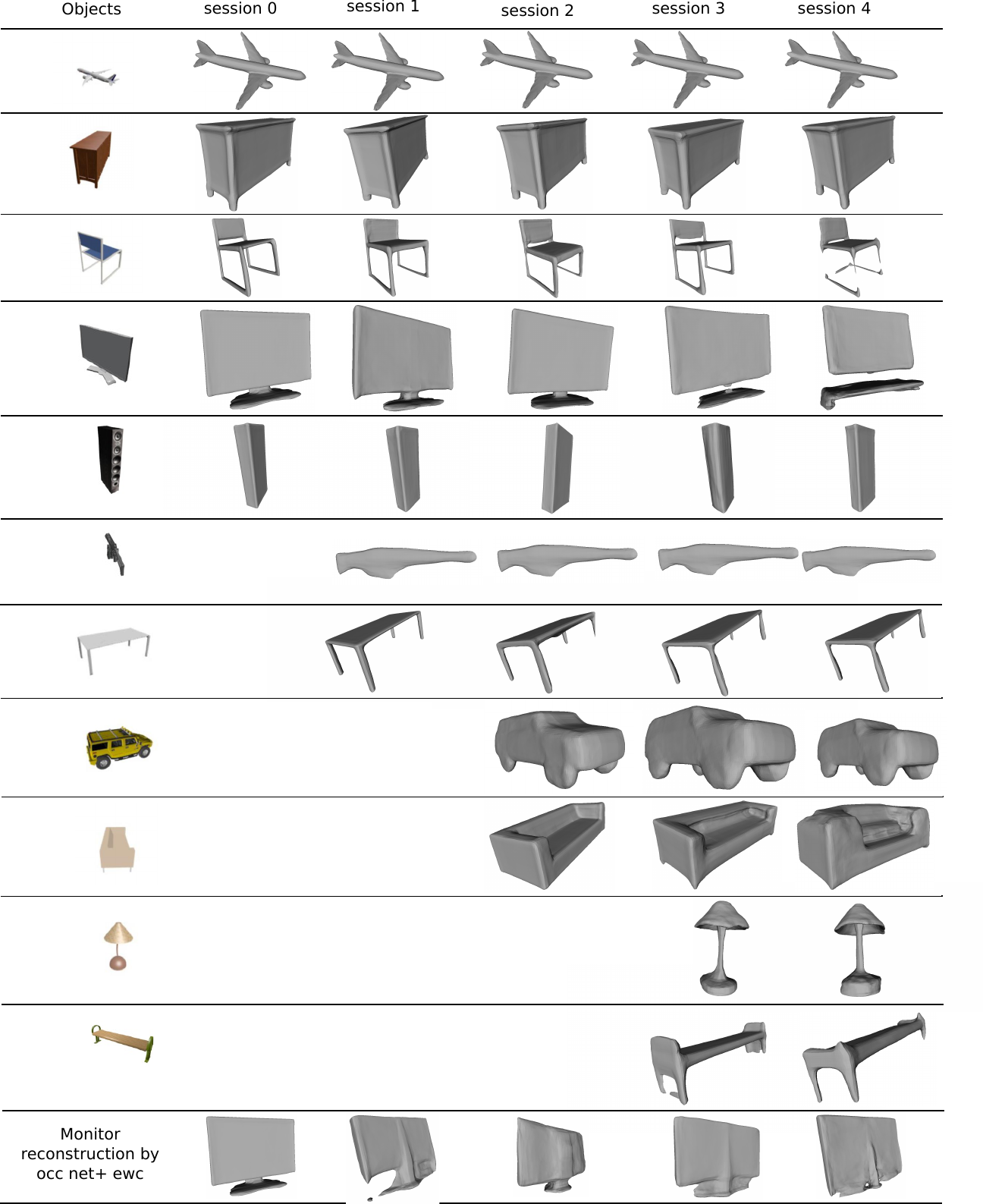}
        \caption{\footnotesize{Reconstruction of different objects across 5 sessions of ShapeNet 13 objects.}}
        \label{fig:incre_reconstruct_shapenet_fig}
    \end{minipage}%
    \hspace{0.5mm}
    \hfill
    \begin{minipage}{0.27\textwidth}%
            \centering
                \includegraphics[width=3.7cm, height=3.7cm]{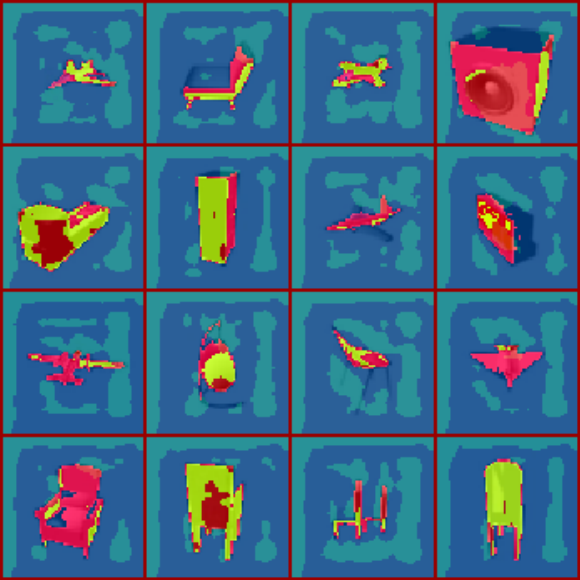}
                \caption{\footnotesize{The saliency maps for various objects acquired during session 0.}} 
                \label{fig:saliency_map}
            \centering
                \includegraphics[width=3.8cm, height=3.8cm]{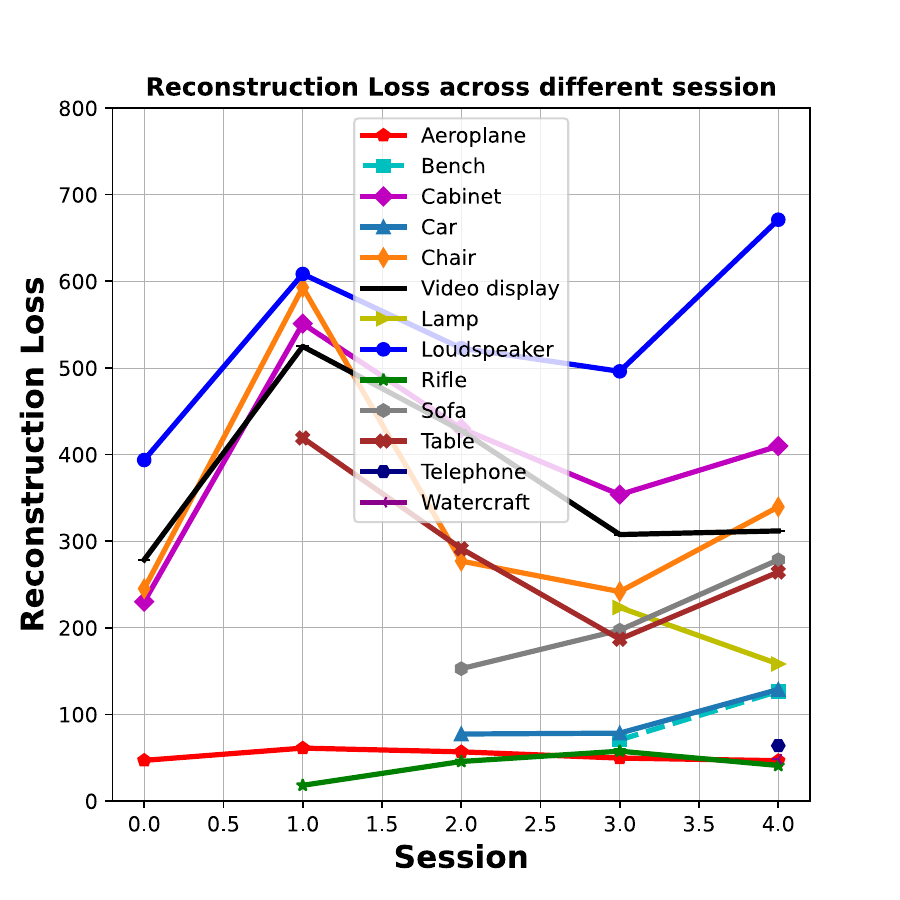}
                \caption{\footnotesize{Reconstruction loss variation across different sessions.}} 
                \label{fig:recon-a}
        \centering
            \includegraphics[width=3.8cm, height=3.8cm]{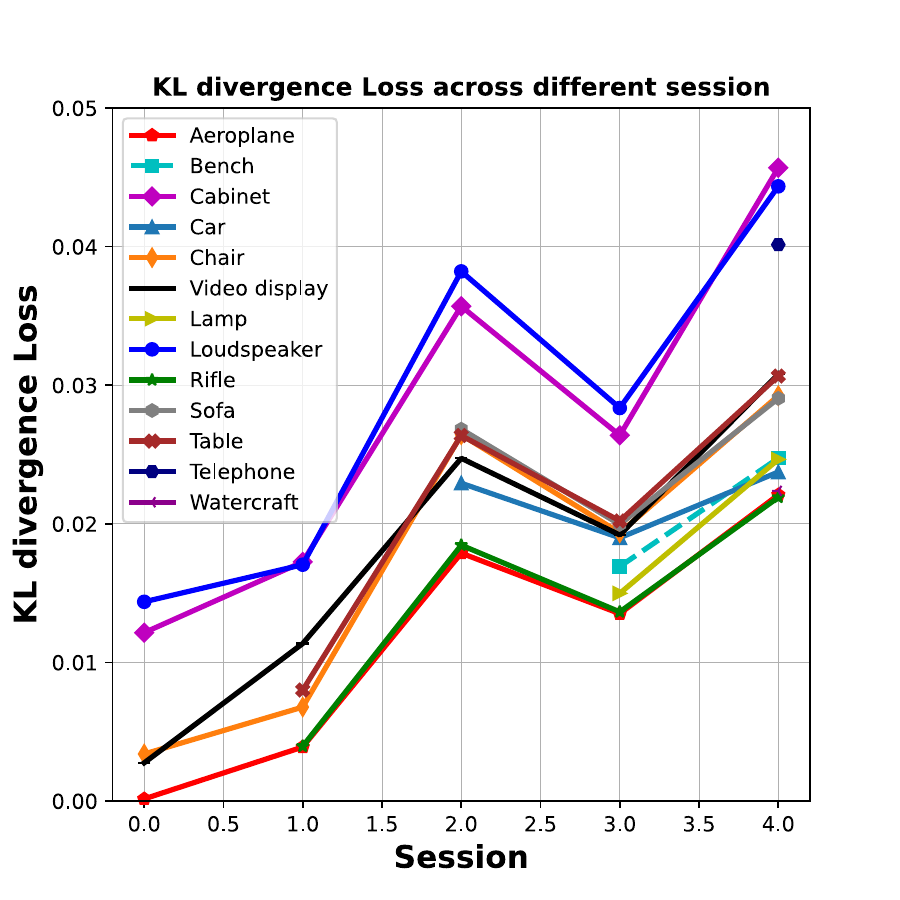}
            \caption{\footnotesize{KL divergence loss variation across different sessions.}} 
            \label{fig:kl-b}
    \end{minipage}
\end{figure*}

\begin{table*}
        \begin{tabular}{c c c c c c c}
        \hline
        \textbf{(a) ShapeNet}&\multicolumn{5}{c}{\textbf{Session}}\\
        \cline{2-6} 
        \textbf{Object} & \textbf{0}& \textbf{1}& \textbf{2}& \textbf{3}& \textbf{4} & \textbf{mean}$\uparrow$ \\
        \hline
        Loudspeaker & \textbf{0.662}
         & 0.637 & 0.645 & 0.614 & 0.596 \\
        Display & \textbf{0.471}
         & 0.446 & 0.463 & 0.451 & 0.446 \\
        Airplane & \textbf{0.571} & 0.559 & 0.565 & 0.552 & 0.547 & \textbf{0.555}\\
        Cabinet & \textbf{0.733}
         & 0.718 & 0.721 & 0.716 & 0.703 \\
        Chair & \textbf{0.501}
         & 0.486 & 0.494 & 0.487 & 0.476 \\
        \hline
        Rifle & - 
         & \textbf{0.486} & 0.471 & 0.452 & 0.441 \\
        Table & -
         & \textbf{0.506} & 0.497 & 0.484 & 0.482 & \textbf{0.462}\\
        \hline
        Car & -
         & - & \textbf{0.742} & 0.723 & 0.731 \\
        Sofa 
         & -
         & - & \textbf{0.681} & 0.670 & 0.666 & \textbf{0.699}\\
        \hline
        Bench & -
         & - & - & \textbf{0.485} & 0.472 \\
        Lamp & -
         & - & - & \textbf{0.371} & 0.367 & \textbf{0.42}\\
        \hline
        Telephone 
        & -
         & - & - & - & \textbf{0.719} \\
        Vessel & -
         & - & - & - & \textbf{0.527} & \textbf{0.623}\\
         \hline
        \end{tabular}
        \hfill
        \begin{tabular}{c c c c c c }
        \hline
        \textbf{(b) KITTI}&\multicolumn{4}{c}{\textbf{Session}}\\
        \cline{2-5} 
        \textbf{Object} & \textbf{0}& \textbf{1}& \textbf{2}& \textbf{3}& \textbf{mean}$\uparrow$ \\
        \hline
        Building & \textbf{0.742}
         & 0.735 & 0.728 & 0.719 \\
        Vegetation & \textbf{0.471}
         & 0.446 & 0.463 & 0.451 \\
        Car & \textbf{0.721} & 0.713 & 0.705 & 0.702 & \textbf{0.597}\\
        Pedestrian & \textbf{0.597}
         & 0.518 & 0.521 & 0.516\\
        \hline
         Cyclist &\textbf{0.601}
         & 0.586 & 0.594 & 0.587& \textbf{0.675}\\
        Sign & - 
         & \textbf{0.786} & 0.771 & 0.762\\
        \hline
        Fence & -
         & \textbf{0.542} & 0.527 & 0.531 & \textbf{0.577}\\
        Tree & -
         & - & \textbf{0.637} & 0.623  \\
        \hline
        Pole 
         & -
         & - & \textbf{0.781} & 0.770 &  \textbf{0.678}\\
        Sidewalk & -
         & - & - & \textbf{0.585}  \\
        \hline
        \label{table:IOU_var_kitti}
        \end{tabular}
    \caption{\footnotesize{(a)Variation in Intersection over Union (IOU) values for distinct objects in the ShapeNet dataset across multiple sessions ($i$-IOU=0.551).
    (b) Variation in Intersection over Union (IOU) values for distinct objects in the KITTI dataset across multiple sessions($i$-IOU=0.631).}}
     \label{table:IOU_var_shapenet_kitti}
\end{table*}

\begin{table*}
        \begin{tabular}{l  l  l}
        \textbf{(a) Method} & \textbf{$i$-IOU}$\uparrow$ & \textbf{BWT}$\uparrow$ \\
        \hline \hline
        Occ\cite{mescheder2019occupancy}(Joint) & 0.588 & -\\
        \hline
        Occ\cite{mescheder2019occupancy}+EWC\cite{kirkpatrick2017overcoming} & 0.532 & -0.056\\
        Occ\cite{mescheder2019occupancy}+GEM\cite{lopez2017gradient} & 0.521 & -0.067\\
        Occ\cite{mescheder2019occupancy}+DEN\cite{yoon2017lifelong} & 0.517 & -0.071\\
        SDF\cite{thai20213d}+EWC\cite{kirkpatrick2017overcoming} & 0.542 & -0.047\\
        SDF\cite{thai20213d}+DEN\cite{yoon2017lifelong} & 0.521 & -0.067\\
        SDF\cite{thai20213d}+GEM\cite{lopez2017gradient} & 0.535 & -0.059\\
        \textbf{CVDP (Ours)} & \textbf{0.551} & \textbf{-0.034}\\
        \hline
        \end{tabular}
        \begin{tabular}{l  l  l}
        \textbf{(b) Saliency Method} & \textbf{$i$-IOU}$\uparrow$ & \textbf{BWT}$\uparrow$\\
        \hline \hline
        Grad-Cam\cite{selvaraju2017grad} & 0.551 & -0.034\\
        Full-Grad\cite{srinivas2019full} & 0.521 & -0.067\\
        Smooth-Grad\cite{smilkov2017smoothgrad} & 0.521 & -0.067\\
        LPA\cite{jetley2018learn} & 0.532 & -0.056\\
        \hline
        \end{tabular}
        \begin{tabular}{l  l  l}
        \textbf{(c)Local saliency insertion} & \textbf{$i$-IOU}$\uparrow$ & \textbf{BWT}$\uparrow$ \\
        \hline \hline
        \textbf{CVDP (\textsc{Exact})} & \textbf{0.551} & \textbf{-0.034}\\
        CVDP (\textsc{Zero-Pad})& 0.521 & -0.067\\
        CVDP (\textsc{Comp and Int}) & 0.532 & -0.056\\
        \hline
        \textsc{Random Patch Replay} & 0.510 & -0.078 \\
        \textsc{Compressed replay} & 0.527 & -0.061 \\
        \hline
        \end{tabular}
     \caption{\footnotesize{(a) Comparison of single-image 3D reconstruction performance on ShapeNet using various methods is conducted by assessing the $i$-IOU.(b) The influence of various saliency map generation methods on ShapeNet 13 dataset is assessed. (c) Effect of diverse image augmentation and patch-based image generation techniques}}
     \label{tab:method_comp_sal_map_pad_ablation}
\end{table*}



\textbf{Metrics}
We employ Intersection over Union (IOU) as our evaluation metric, specifically examining the incremental IOU of the models across various sessions. The IOU is computed using the following formula:
$$
\mathrm{IOU} = \frac{\sum_{i,j,k}\mathbb{I}(\hat{\mathrm{p}}_{i,j,k}>\mathrm{t})\mathbb{I}(\mathrm{p}_{i,j,k})}{\sum_{i,j,k}\mathbb{I}(\mathbb{I}(\hat{\mathrm{p}}_{i,j,k}>\mathrm{t})+\mathbb{I}(\mathrm{p}_{i,j,k}))}
$$
In this context, the indicator function $\mathbb{I}(.)$ outputs values of zero or one. $\hat{\mathrm{p}}_{i,j,k}$ and $\mathrm{p}_{i,j,k}$ represent the predicted probability and the ground truth, respectively. The threshold $t$ is fixed at 0.2 in our experiments.

\textbf{Incremental-IOU and Backward Transfer:}
The incremental Intersection Over Union ($i$-IOU) and Backward Transfer (BWT), within the context of incremental learning, respectively indicate the model's performance on test data and its improvement in performance on the previous data. The $i$-IOU (average accuracy) loss for an object is defined as follows:
$$
i-\mathrm{IOU} = \frac{1}{\mathrm{T}} \sum_{j=1}^{\mathrm{T}}\sum_{k=1}^{\mathcal{N}_t} \mathrm{IOU}_{\mathrm{T}, j}^{k}
$$
and Backward Transfer(BWT) for a session is given as,
$$
\mathrm{BWT} = \frac{1}{\mathrm{T}-1}\sum_{i=1}^{ \mathrm{T}-1}\sum_{k=1}^{\mathcal{N}_t} \mathrm{IOU}_{\mathrm{T},i}^{k} - \mathrm{IOU}_{i,i}^{k}
$$
In this context, $\mathrm{IOU}_{i,j}^{k}$ represents the Intersection over Union (IOU) of $k$-th object at session $j$ after the model has been trained on session $i$. After training the model on the current set of objects, we assess its performance on all previous tasks. This process involves constructing an upper triangular matrix, where each column represents a specific time step and each row corresponds to a task number. Negative BWT signifies catastrophic forgetting, and our ideal goal is to achieve a BWT value of zero, which indicates the absence of catastrophic forgetting.  


\subsection{Result and Analyses:}
Initially, we compute the IOU values for various objects across different sessions and calculate the mean IOU for the objects in each session after the final session for the ShapeNet and KITTI datasets. These results are detailed in Table~\ref{table:IOU_var_shapenet_kitti}(a) and Table~\ref{table:IOU_var_shapenet_kitti}(b). The IOU values are calculated with different numbers of memory blocks, $n_{\mathcal{B}}$, available per class for the saliency maps. These memory blocks serve the purpose of storing saliency maps for points belonging to different class objects, along with their associated ground truth. As a result, the size of the memory is expressed as $ |\mathcal{M}_{\mathrm{SAL}}| = n_{\mathcal{B}}\cdot \Sigma_{t=1}^{T} \mathcal{N}_t$.
Each saliency map pertains to a specific region within an object's image and is represented as a pixel matrix of dimensions $H \times W$. Each entry in the memory module also includes occupancy information associated with the respective object. This binary value indicates voxel occupation and is represented by $R^3$ parameters, considering the voxel resolution as $R$. For practicality and to adhere to voxel budget constraints, we sample and store 1024 points per object. Therefore $n_{\mathcal{B}} = k \times (H \times W) + R^3$ with $k$ number of saliency (local and global) per object.
Finally, we report the $i$-IOU for our method, CVDP (Combining Variational Distribution Prior), and compare it with other baselines on the ShapeNet dataset, as detailed in Table~\ref{tab:method_comp_sal_map_pad_ablation}(a). Additionally, we provide qualitative results across different sessions in Figure~\ref{fig:incre_reconstruct_shapenet_fig}.

\begin{figure}[!h]
  \centering
   \includegraphics[width=1.0\linewidth]{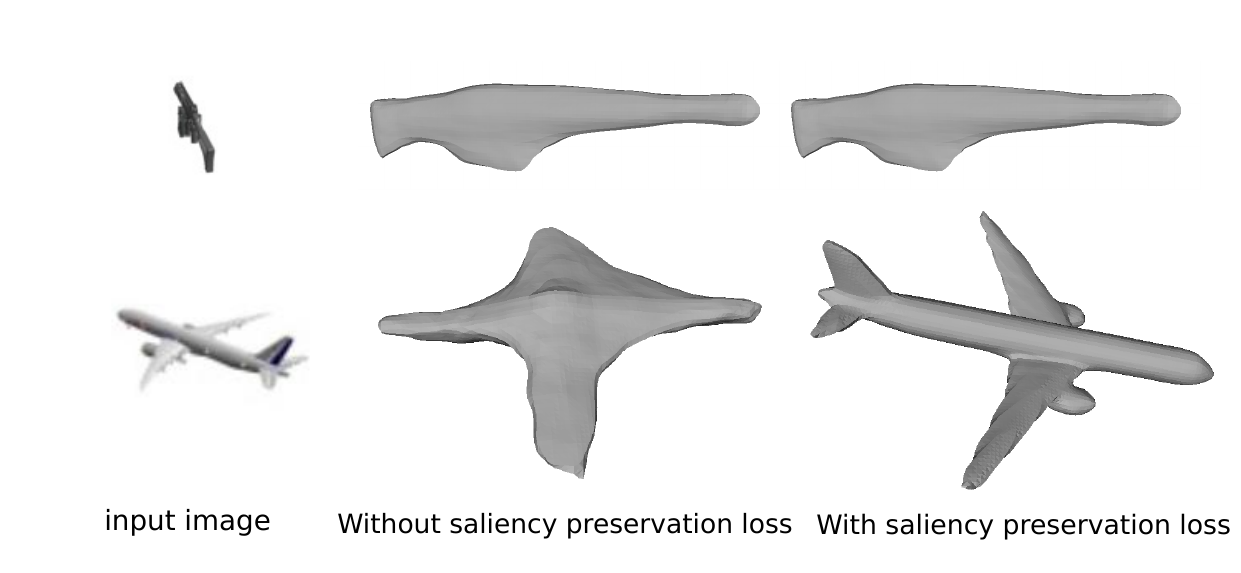}
   \caption{\footnotesize{Without saliency preservation, the model experiences a loss of the distinctive attributes associated with the "airplane" (session 0) class after being trained on "rifle" (session 1).}}
   \label{fig:saliency_ablation}
\end{figure}

\textbf{Latent codes of different objects:}
We employed t-SNE \cite{van2008visualizing} dimensionality reduction to visualize the mean latent distribution of various objects. Figure \ref{fig:latent_mean} exhibit the latent codes of distinct objects of ShapeNet. In the initial session (session 0), it's evident that the latent code for the "airplane" object minimally overlaps with the latent codes of other shapes within the same session. Consequently, the "airplane" object is more susceptible to forgetting as the network is trained on subsequent sessions. This underscores the importance of local and global image saliency maps in retaining the knowledge of previous session objects. When the model is trained without image saliency, it becomes more attuned to the "rifle" object during session 1, thereby causing a loss of significant distinctive information related to the "airplane" (Figure~\ref{fig:saliency_ablation}). 

\textbf{KL divergence and reconstruction loss:}
The KL divergence loss provides a measure of the disparity between the latent code distribution of the current session's object and the approximated cumulative variational prior. This demonstrates the ability of the variational priors to retain past knowledge and to generate novel objects. As illustrated in Fig.~\ref{fig:kl-b}, the KL divergence loss for all objects remains consistently below 0.05. An escalation in the KL divergence loss can be attributed to the aggregation of variational priors, which contains a smaller fraction of priors corresponding to specific objects. Similarly, the reconstruction loss is illustrated in Fig.~\ref{fig:recon-a}
Observing the figure, it becomes evident that the reconstruction loss for the "table" object has shown a substantial decrease as we transition to different sessions, signifying a positive transfer of knowledge for this particular object.
\textbf{Backward Transfer:}
During the training of the model in session 2, a notable amount of Backward transfer is observed. Specifically, when training the model for the "car" category, it effectively extends its acquired knowledge to the base session. This phenomenon occurs due to the model learning distinctive features that contribute to enhancing the knowledge for the base session. This is illustrated in Table~\ref{table:IOU_var_shapenet_kitti}(a). This can be assessed by examining the mean values of objects generated in session 0 (first 5 rows of columns 0, 1, and 2). Specifically, the values for sessions 0, 1, and 2 are 0.588, 0.57, and 0.578 respectively. Therefore, an increase of 0.008 is observed from session 1 to session 2.
\subsection{Ablation experiments:}
\label{sec:ablation}

\textbf{Image saliency:}
The inclusion of the image saliency loss holds significance in maintaining both local and global distinctive attributes across various shapes. For instance, during session 0, the model is trained on the "Aeroplane" class. However, in session 1, the training focuses on the "Gun" class. Without the preservation of image saliency, the model could lose some distinctive features of the "Aeroplane" class, such as its cylindrical shape, engine, and the wing's length. This phenomenon is depicted in Fig.~\ref{fig:saliency_ablation}. Various techniques have been employed to generate saliency maps for objects, which are detailed in Table~\ref{tab:method_comp_sal_map_pad_ablation}(b). We have conducted a comparison with GRAD-CAM \cite{selvaraju2017grad}, smooth-grad \cite{smilkov2017smoothgrad}, full-grad \cite{srinivas2019full}, and the learn to pay attention module \cite{jetley2018learn} to determine the significant regions of the image or the cropped image.


\textbf{Padding and interpolation of patches:}
Three strategies have been employed to determine the arrangement of cropped patches during the image regeneration process.  The effects of these methodologies are illustrated in Table~\ref{tab:method_comp_sal_map_pad_ablation}(c). The first approach involves placing all patches while preserving their coordinates (CVDP (\textsc{Exact})). Alternatively, patches can be zero-padded to attain the desired shape (CVDP ((\textsc{Zero-Pad})). Another option includes compressing the image to a lower resolution and storing the resulting compressed saliency map. During rehearsal, the image is then interpolated back to its original form (CVDP (\textsc{Comp and Int})). Additionally, results are provided for randomly replaying patches and compressed replay.


\section{Conclusion:}
In this study, we systematically demonstrate the feasibility of training a single-image 3D reconstruction model within the Continual Learning paradigm. We then showcase an efficient strategy to counteract catastrophic forgetting in this framework, employing minimal memory resources. Moreover, we introduce an approach for judiciously storing 3D shapes, enabling the model to harness this knowledge effectively for various tasks amidst evolving environments. It is our aspiration that this endeavor will inaugurate novel opportunities in the realm of single-image 3D reconstruction.

\bibliographystyle{ACM-Reference-Format}
\bibliography{ICVGIP-Latex-Template}


\begin{thebibliography}{62}


\ifx \showCODEN    \undefined \def \showCODEN     #1{\unskip}     \fi
\ifx \showDOI      \undefined \def \showDOI       #1{#1}\fi
\ifx \showISBNx    \undefined \def \showISBNx     #1{\unskip}     \fi
\ifx \showISBNxiii \undefined \def \showISBNxiii  #1{\unskip}     \fi
\ifx \showISSN     \undefined \def \showISSN      #1{\unskip}     \fi
\ifx \showLCCN     \undefined \def \showLCCN      #1{\unskip}     \fi
\ifx \shownote     \undefined \def \shownote      #1{#1}          \fi
\ifx \showarticletitle \undefined \def \showarticletitle #1{#1}   \fi
\ifx \showURL      \undefined \def \showURL       {\relax}        \fi
\providecommand\bibfield[2]{#2}
\providecommand\bibinfo[2]{#2}
\providecommand\natexlab[1]{#1}
\providecommand\showeprint[2][]{arXiv:#2}

\bibitem[\protect\citeauthoryear{Abel, Arumugam, Lehnert, and Littman}{Abel et~al\mbox{.}}{2018}]%
        {abel2018state}
\bibfield{author}{\bibinfo{person}{David Abel}, \bibinfo{person}{Dilip Arumugam}, \bibinfo{person}{Lucas Lehnert}, {and} \bibinfo{person}{Michael Littman}.} \bibinfo{year}{2018}\natexlab{}.
\newblock \showarticletitle{State abstractions for lifelong reinforcement learning}. In \bibinfo{booktitle}{\emph{International Conference on Machine Learning}}. PMLR, \bibinfo{pages}{10--19}.
\newblock


\bibitem[\protect\citeauthoryear{Aljundi, Kelchtermans, and Tuytelaars}{Aljundi et~al\mbox{.}}{2019a}]%
        {aljundi2019task}
\bibfield{author}{\bibinfo{person}{Rahaf Aljundi}, \bibinfo{person}{Klaas Kelchtermans}, {and} \bibinfo{person}{Tinne Tuytelaars}.} \bibinfo{year}{2019}\natexlab{a}.
\newblock \showarticletitle{Task-free continual learning}. In \bibinfo{booktitle}{\emph{Proceedings of the IEEE/CVF Conference on Computer Vision and Pattern Recognition}}. \bibinfo{pages}{11254--11263}.
\newblock


\bibitem[\protect\citeauthoryear{Aljundi, Lin, Goujaud, and Bengio}{Aljundi et~al\mbox{.}}{2019b}]%
        {aljundi2019gradient}
\bibfield{author}{\bibinfo{person}{Rahaf Aljundi}, \bibinfo{person}{Min Lin}, \bibinfo{person}{Baptiste Goujaud}, {and} \bibinfo{person}{Yoshua Bengio}.} \bibinfo{year}{2019}\natexlab{b}.
\newblock \showarticletitle{Gradient based sample selection for online continual learning}.
\newblock \bibinfo{journal}{\emph{Advances in neural information processing systems}}  \bibinfo{volume}{32} (\bibinfo{year}{2019}).
\newblock


\bibitem[\protect\citeauthoryear{Bozic, Zollhofer, Theobalt, and Nie{\ss}ner}{Bozic et~al\mbox{.}}{2020}]%
        {bozic2020deepdeform}
\bibfield{author}{\bibinfo{person}{Aljaz Bozic}, \bibinfo{person}{Michael Zollhofer}, \bibinfo{person}{Christian Theobalt}, {and} \bibinfo{person}{Matthias Nie{\ss}ner}.} \bibinfo{year}{2020}\natexlab{}.
\newblock \showarticletitle{Deepdeform: Learning non-rigid rgb-d reconstruction with semi-supervised data}. In \bibinfo{booktitle}{\emph{Proceedings of the IEEE/CVF Conference on Computer Vision and Pattern Recognition}}. \bibinfo{pages}{7002--7012}.
\newblock


\bibitem[\protect\citeauthoryear{Cadena, Carlone, Carrillo, Latif, Scaramuzza, Neira, Reid, and Leonard}{Cadena et~al\mbox{.}}{2016}]%
        {cadena2016past}
\bibfield{author}{\bibinfo{person}{Cesar Cadena}, \bibinfo{person}{Luca Carlone}, \bibinfo{person}{Henry Carrillo}, \bibinfo{person}{Yasir Latif}, \bibinfo{person}{Davide Scaramuzza}, \bibinfo{person}{Jos{\'e} Neira}, \bibinfo{person}{Ian Reid}, {and} \bibinfo{person}{John~J Leonard}.} \bibinfo{year}{2016}\natexlab{}.
\newblock \showarticletitle{Past, present, and future of simultaneous localization and mapping: Toward the robust-perception age}.
\newblock \bibinfo{journal}{\emph{IEEE Transactions on robotics}} \bibinfo{volume}{32}, \bibinfo{number}{6} (\bibinfo{year}{2016}), \bibinfo{pages}{1309--1332}.
\newblock


\bibitem[\protect\citeauthoryear{Cermelli, Mancini, Bulo, Ricci, and Caputo}{Cermelli et~al\mbox{.}}{2020}]%
        {cermelli2020modeling}
\bibfield{author}{\bibinfo{person}{Fabio Cermelli}, \bibinfo{person}{Massimiliano Mancini}, \bibinfo{person}{Samuel~Rota Bulo}, \bibinfo{person}{Elisa Ricci}, {and} \bibinfo{person}{Barbara Caputo}.} \bibinfo{year}{2020}\natexlab{}.
\newblock \showarticletitle{Modeling the background for incremental learning in semantic segmentation}. In \bibinfo{booktitle}{\emph{Proceedings of the IEEE/CVF Conference on Computer Vision and Pattern Recognition}}. \bibinfo{pages}{9233--9242}.
\newblock


\bibitem[\protect\citeauthoryear{Chaudhry, Rohrbach, Elhoseiny, Ajanthan, Dokania, Torr, and Ranzato}{Chaudhry et~al\mbox{.}}{2019}]%
        {chaudhry2019continual}
\bibfield{author}{\bibinfo{person}{Arslan Chaudhry}, \bibinfo{person}{Marcus Rohrbach}, \bibinfo{person}{Mohamed Elhoseiny}, \bibinfo{person}{Thalaiyasingam Ajanthan}, \bibinfo{person}{P Dokania}, \bibinfo{person}{P Torr}, {and} \bibinfo{person}{M Ranzato}.} \bibinfo{year}{2019}\natexlab{}.
\newblock \showarticletitle{Continual learning with tiny episodic memories}. In \bibinfo{booktitle}{\emph{Workshop on Multi-Task and Lifelong Reinforcement Learning}}.
\newblock


\bibitem[\protect\citeauthoryear{Choy, Xu, Gwak, Chen, and Savarese}{Choy et~al\mbox{.}}{2016}]%
        {choy20163d}
\bibfield{author}{\bibinfo{person}{Christopher~B Choy}, \bibinfo{person}{Danfei Xu}, \bibinfo{person}{JunYoung Gwak}, \bibinfo{person}{Kevin Chen}, {and} \bibinfo{person}{Silvio Savarese}.} \bibinfo{year}{2016}\natexlab{}.
\newblock \showarticletitle{3d-r2n2: A unified approach for single and multi-view 3d object reconstruction}. In \bibinfo{booktitle}{\emph{Computer Vision--ECCV 2016: 14th European Conference, Amsterdam, The Netherlands, October 11-14, 2016, Proceedings, Part VIII 14}}. Springer, \bibinfo{pages}{628--644}.
\newblock


\bibitem[\protect\citeauthoryear{Geiger, Lenz, Stiller, and Urtasun}{Geiger et~al\mbox{.}}{2013}]%
        {geiger2013vision}
\bibfield{author}{\bibinfo{person}{Andreas Geiger}, \bibinfo{person}{Philip Lenz}, \bibinfo{person}{Christoph Stiller}, {and} \bibinfo{person}{Raquel Urtasun}.} \bibinfo{year}{2013}\natexlab{}.
\newblock \showarticletitle{Vision meets robotics: The kitti dataset}.
\newblock \bibinfo{journal}{\emph{The International Journal of Robotics Research}} \bibinfo{volume}{32}, \bibinfo{number}{11} (\bibinfo{year}{2013}), \bibinfo{pages}{1231--1237}.
\newblock


\bibitem[\protect\citeauthoryear{Hayes, Cahill, and Kanan}{Hayes et~al\mbox{.}}{2019}]%
        {hayes2019memory}
\bibfield{author}{\bibinfo{person}{Tyler~L Hayes}, \bibinfo{person}{Nathan~D Cahill}, {and} \bibinfo{person}{Christopher Kanan}.} \bibinfo{year}{2019}\natexlab{}.
\newblock \showarticletitle{Memory efficient experience replay for streaming learning}. In \bibinfo{booktitle}{\emph{2019 International Conference on Robotics and Automation (ICRA)}}. IEEE, \bibinfo{pages}{9769--9776}.
\newblock


\bibitem[\protect\citeauthoryear{Hern{\'a}ndez-Lobato and Adams}{Hern{\'a}ndez-Lobato and Adams}{2015}]%
        {hernandez2015probabilistic}
\bibfield{author}{\bibinfo{person}{Jos{\'e}~Miguel Hern{\'a}ndez-Lobato} {and} \bibinfo{person}{Ryan Adams}.} \bibinfo{year}{2015}\natexlab{}.
\newblock \showarticletitle{Probabilistic backpropagation for scalable learning of bayesian neural networks}. In \bibinfo{booktitle}{\emph{International conference on machine learning}}. PMLR, \bibinfo{pages}{1861--1869}.
\newblock


\bibitem[\protect\citeauthoryear{Jetley, Lord, Lee, and Torr}{Jetley et~al\mbox{.}}{2018}]%
        {jetley2018learn}
\bibfield{author}{\bibinfo{person}{Saumya Jetley}, \bibinfo{person}{Nicholas~A Lord}, \bibinfo{person}{Namhoon Lee}, {and} \bibinfo{person}{Philip~HS Torr}.} \bibinfo{year}{2018}\natexlab{}.
\newblock \showarticletitle{Learn to pay attention}.
\newblock \bibinfo{journal}{\emph{arXiv preprint arXiv:1804.02391}} (\bibinfo{year}{2018}).
\newblock


\bibitem[\protect\citeauthoryear{Jimenez~Rezende, Eslami, Mohamed, Battaglia, Jaderberg, and Heess}{Jimenez~Rezende et~al\mbox{.}}{2016}]%
        {jimenez2016unsupervised}
\bibfield{author}{\bibinfo{person}{Danilo Jimenez~Rezende}, \bibinfo{person}{SM Eslami}, \bibinfo{person}{Shakir Mohamed}, \bibinfo{person}{Peter Battaglia}, \bibinfo{person}{Max Jaderberg}, {and} \bibinfo{person}{Nicolas Heess}.} \bibinfo{year}{2016}\natexlab{}.
\newblock \showarticletitle{Unsupervised learning of 3d structure from images}.
\newblock \bibinfo{journal}{\emph{Advances in neural information processing systems}}  \bibinfo{volume}{29} (\bibinfo{year}{2016}).
\newblock


\bibitem[\protect\citeauthoryear{Kirkpatrick, Pascanu, Rabinowitz, Veness, Desjardins, Rusu, Milan, Quan, Ramalho, Grabska-Barwinska, et~al\mbox{.}}{Kirkpatrick et~al\mbox{.}}{2017}]%
        {kirkpatrick2017overcoming}
\bibfield{author}{\bibinfo{person}{James Kirkpatrick}, \bibinfo{person}{Razvan Pascanu}, \bibinfo{person}{Neil Rabinowitz}, \bibinfo{person}{Joel Veness}, \bibinfo{person}{Guillaume Desjardins}, \bibinfo{person}{Andrei~A Rusu}, \bibinfo{person}{Kieran Milan}, \bibinfo{person}{John Quan}, \bibinfo{person}{Tiago Ramalho}, \bibinfo{person}{Agnieszka Grabska-Barwinska}, {et~al\mbox{.}}} \bibinfo{year}{2017}\natexlab{}.
\newblock \showarticletitle{Overcoming catastrophic forgetting in neural networks}.
\newblock \bibinfo{journal}{\emph{Proceedings of the national academy of sciences}} \bibinfo{volume}{114}, \bibinfo{number}{13} (\bibinfo{year}{2017}), \bibinfo{pages}{3521--3526}.
\newblock


\bibitem[\protect\citeauthoryear{Lesort, Caselles-Dupr{\'e}, Garcia-Ortiz, Stoian, and Filliat}{Lesort et~al\mbox{.}}{2019}]%
        {lesort2019generative}
\bibfield{author}{\bibinfo{person}{Timoth{\'e}e Lesort}, \bibinfo{person}{Hugo Caselles-Dupr{\'e}}, \bibinfo{person}{Michael Garcia-Ortiz}, \bibinfo{person}{Andrei Stoian}, {and} \bibinfo{person}{David Filliat}.} \bibinfo{year}{2019}\natexlab{}.
\newblock \showarticletitle{Generative models from the perspective of continual learning}. In \bibinfo{booktitle}{\emph{2019 International Joint Conference on Neural Networks (IJCNN)}}. IEEE, \bibinfo{pages}{1--8}.
\newblock


\bibitem[\protect\citeauthoryear{Liu, Wu, Van~Hoorick, Tokmakov, Zakharov, and Vondrick}{Liu et~al\mbox{.}}{2023}]%
        {liu2023zero}
\bibfield{author}{\bibinfo{person}{Ruoshi Liu}, \bibinfo{person}{Rundi Wu}, \bibinfo{person}{Basile Van~Hoorick}, \bibinfo{person}{Pavel Tokmakov}, \bibinfo{person}{Sergey Zakharov}, {and} \bibinfo{person}{Carl Vondrick}.} \bibinfo{year}{2023}\natexlab{}.
\newblock \showarticletitle{Zero-1-to-3: Zero-shot one image to 3d object}.
\newblock \bibinfo{journal}{\emph{arXiv preprint arXiv:2303.11328}} (\bibinfo{year}{2023}).
\newblock


\bibitem[\protect\citeauthoryear{Lopez-Paz and Ranzato}{Lopez-Paz and Ranzato}{2017}]%
        {lopez2017gradient}
\bibfield{author}{\bibinfo{person}{David Lopez-Paz} {and} \bibinfo{person}{Marc'Aurelio Ranzato}.} \bibinfo{year}{2017}\natexlab{}.
\newblock \showarticletitle{Gradient episodic memory for continual learning}.
\newblock \bibinfo{journal}{\emph{Advances in neural information processing systems}}  \bibinfo{volume}{30} (\bibinfo{year}{2017}).
\newblock


\bibitem[\protect\citeauthoryear{Maracani, Michieli, Toldo, and Zanuttigh}{Maracani et~al\mbox{.}}{2021}]%
        {maracani2021recall}
\bibfield{author}{\bibinfo{person}{Andrea Maracani}, \bibinfo{person}{Umberto Michieli}, \bibinfo{person}{Marco Toldo}, {and} \bibinfo{person}{Pietro Zanuttigh}.} \bibinfo{year}{2021}\natexlab{}.
\newblock \showarticletitle{Recall: Replay-based continual learning in semantic segmentation}. In \bibinfo{booktitle}{\emph{Proceedings of the IEEE/CVF international conference on computer vision}}. \bibinfo{pages}{7026--7035}.
\newblock


\bibitem[\protect\citeauthoryear{Marton, Pangercic, Blodow, Kleinehellefort, and Beetz}{Marton et~al\mbox{.}}{2010}]%
        {marton2010general}
\bibfield{author}{\bibinfo{person}{Zoltan-Csaba Marton}, \bibinfo{person}{Dejan Pangercic}, \bibinfo{person}{Nico Blodow}, \bibinfo{person}{Jonathan Kleinehellefort}, {and} \bibinfo{person}{Michael Beetz}.} \bibinfo{year}{2010}\natexlab{}.
\newblock \showarticletitle{General 3D modelling of novel objects from a single view}. In \bibinfo{booktitle}{\emph{2010 ieee/rsj international conference on intelligent robots and systems}}. IEEE, \bibinfo{pages}{3700--3705}.
\newblock


\bibitem[\protect\citeauthoryear{Maturana and Scherer}{Maturana and Scherer}{2015}]%
        {maturana2015voxnet}
\bibfield{author}{\bibinfo{person}{Daniel Maturana} {and} \bibinfo{person}{Sebastian Scherer}.} \bibinfo{year}{2015}\natexlab{}.
\newblock \showarticletitle{Voxnet: A 3d convolutional neural network for real-time object recognition}. In \bibinfo{booktitle}{\emph{2015 IEEE/RSJ international conference on intelligent robots and systems (IROS)}}. IEEE, \bibinfo{pages}{922--928}.
\newblock


\bibitem[\protect\citeauthoryear{Mescheder, Oechsle, Niemeyer, Nowozin, and Geiger}{Mescheder et~al\mbox{.}}{2019}]%
        {mescheder2019occupancy}
\bibfield{author}{\bibinfo{person}{Lars Mescheder}, \bibinfo{person}{Michael Oechsle}, \bibinfo{person}{Michael Niemeyer}, \bibinfo{person}{Sebastian Nowozin}, {and} \bibinfo{person}{Andreas Geiger}.} \bibinfo{year}{2019}\natexlab{}.
\newblock \showarticletitle{Occupancy networks: Learning 3d reconstruction in function space}. In \bibinfo{booktitle}{\emph{Proceedings of the IEEE/CVF conference on computer vision and pattern recognition}}. \bibinfo{pages}{4460--4470}.
\newblock


\bibitem[\protect\citeauthoryear{Michalkiewicz, Parisot, Tsogkas, Baktashmotlagh, Eriksson, and Belilovsky}{Michalkiewicz et~al\mbox{.}}{2020}]%
        {michalkiewicz2020few}
\bibfield{author}{\bibinfo{person}{Mateusz Michalkiewicz}, \bibinfo{person}{Sarah Parisot}, \bibinfo{person}{Stavros Tsogkas}, \bibinfo{person}{Mahsa Baktashmotlagh}, \bibinfo{person}{Anders Eriksson}, {and} \bibinfo{person}{Eugene Belilovsky}.} \bibinfo{year}{2020}\natexlab{}.
\newblock \showarticletitle{Few-shot single-view 3-d object reconstruction with compositional priors}. In \bibinfo{booktitle}{\emph{European Conference on Computer Vision}}. Springer, \bibinfo{pages}{614--630}.
\newblock


\bibitem[\protect\citeauthoryear{Michieli and Zanuttigh}{Michieli and Zanuttigh}{2019}]%
        {michieli2019incremental}
\bibfield{author}{\bibinfo{person}{Umberto Michieli} {and} \bibinfo{person}{Pietro Zanuttigh}.} \bibinfo{year}{2019}\natexlab{}.
\newblock \showarticletitle{Incremental learning techniques for semantic segmentation}. In \bibinfo{booktitle}{\emph{Proceedings of the IEEE/CVF international conference on computer vision workshops}}. \bibinfo{pages}{0--0}.
\newblock


\bibitem[\protect\citeauthoryear{Nguyen, Li, Bui, and Turner}{Nguyen et~al\mbox{.}}{2017}]%
        {nguyen2017variational}
\bibfield{author}{\bibinfo{person}{Cuong~V Nguyen}, \bibinfo{person}{Yingzhen Li}, \bibinfo{person}{Thang~D Bui}, {and} \bibinfo{person}{Richard~E Turner}.} \bibinfo{year}{2017}\natexlab{}.
\newblock \showarticletitle{Variational continual learning}.
\newblock \bibinfo{journal}{\emph{arXiv preprint arXiv:1710.10628}} (\bibinfo{year}{2017}).
\newblock


\bibitem[\protect\citeauthoryear{Palit, Banerjee, and Chaudhuri}{Palit et~al\mbox{.}}{2022}]%
        {palit2022prototypical}
\bibfield{author}{\bibinfo{person}{Sanchar Palit}, \bibinfo{person}{Biplab Banerjee}, {and} \bibinfo{person}{Subhasis Chaudhuri}.} \bibinfo{year}{2022}\natexlab{}.
\newblock \showarticletitle{Prototypical quadruplet for few-shot class incremental learning}.
\newblock \bibinfo{journal}{\emph{arXiv preprint arXiv:2211.02947}} (\bibinfo{year}{2022}).
\newblock


\bibitem[\protect\citeauthoryear{Rebuffi, Kolesnikov, Sperl, and Lampert}{Rebuffi et~al\mbox{.}}{2017}]%
        {rebuffi2017icarl}
\bibfield{author}{\bibinfo{person}{Sylvestre-Alvise Rebuffi}, \bibinfo{person}{Alexander Kolesnikov}, \bibinfo{person}{Georg Sperl}, {and} \bibinfo{person}{Christoph~H Lampert}.} \bibinfo{year}{2017}\natexlab{}.
\newblock \showarticletitle{icarl: Incremental classifier and representation learning}. In \bibinfo{booktitle}{\emph{Proceedings of the IEEE conference on Computer Vision and Pattern Recognition}}. \bibinfo{pages}{2001--2010}.
\newblock


\bibitem[\protect\citeauthoryear{Riemer, Cases, Ajemian, Liu, Rish, Tu, and Tesauro}{Riemer et~al\mbox{.}}{2018}]%
        {riemer2018learning}
\bibfield{author}{\bibinfo{person}{Matthew Riemer}, \bibinfo{person}{Ignacio Cases}, \bibinfo{person}{Robert Ajemian}, \bibinfo{person}{Miao Liu}, \bibinfo{person}{Irina Rish}, \bibinfo{person}{Yuhai Tu}, {and} \bibinfo{person}{Gerald Tesauro}.} \bibinfo{year}{2018}\natexlab{}.
\newblock \showarticletitle{Learning to learn without forgetting by maximizing transfer and minimizing interference}.
\newblock \bibinfo{journal}{\emph{arXiv preprint arXiv:1810.11910}} (\bibinfo{year}{2018}).
\newblock


\bibitem[\protect\citeauthoryear{Ros, Ramos, Granados, Bakhtiary, Vazquez, and Lopez}{Ros et~al\mbox{.}}{2015}]%
        {ros2015vision}
\bibfield{author}{\bibinfo{person}{German Ros}, \bibinfo{person}{Sebastian Ramos}, \bibinfo{person}{Manuel Granados}, \bibinfo{person}{Amir Bakhtiary}, \bibinfo{person}{David Vazquez}, {and} \bibinfo{person}{Antonio~M Lopez}.} \bibinfo{year}{2015}\natexlab{}.
\newblock \showarticletitle{Vision-based offline-online perception paradigm for autonomous driving}. In \bibinfo{booktitle}{\emph{2015 IEEE Winter Conference on Applications of Computer Vision}}. IEEE, \bibinfo{pages}{231--238}.
\newblock


\bibitem[\protect\citeauthoryear{Schonberger and Frahm}{Schonberger and Frahm}{2016}]%
        {schonberger2016structure}
\bibfield{author}{\bibinfo{person}{Johannes~L Schonberger} {and} \bibinfo{person}{Jan-Michael Frahm}.} \bibinfo{year}{2016}\natexlab{}.
\newblock \showarticletitle{Structure-from-motion revisited}. In \bibinfo{booktitle}{\emph{Proceedings of the IEEE conference on computer vision and pattern recognition}}. \bibinfo{pages}{4104--4113}.
\newblock


\bibitem[\protect\citeauthoryear{Selvaraju, Cogswell, Das, Vedantam, Parikh, and Batra}{Selvaraju et~al\mbox{.}}{2017}]%
        {selvaraju2017grad}
\bibfield{author}{\bibinfo{person}{Ramprasaath~R Selvaraju}, \bibinfo{person}{Michael Cogswell}, \bibinfo{person}{Abhishek Das}, \bibinfo{person}{Ramakrishna Vedantam}, \bibinfo{person}{Devi Parikh}, {and} \bibinfo{person}{Dhruv Batra}.} \bibinfo{year}{2017}\natexlab{}.
\newblock \showarticletitle{Grad-cam: Visual explanations from deep networks via gradient-based localization}. In \bibinfo{booktitle}{\emph{Proceedings of the IEEE international conference on computer vision}}. \bibinfo{pages}{618--626}.
\newblock


\bibitem[\protect\citeauthoryear{Shim, Mai, Jeong, Sanner, Kim, and Jang}{Shim et~al\mbox{.}}{2021}]%
        {shim2021online}
\bibfield{author}{\bibinfo{person}{Dongsub Shim}, \bibinfo{person}{Zheda Mai}, \bibinfo{person}{Jihwan Jeong}, \bibinfo{person}{Scott Sanner}, \bibinfo{person}{Hyunwoo Kim}, {and} \bibinfo{person}{Jongseong Jang}.} \bibinfo{year}{2021}\natexlab{}.
\newblock \showarticletitle{Online class-incremental continual learning with adversarial shapley value}. In \bibinfo{booktitle}{\emph{Proceedings of the AAAI Conference on Artificial Intelligence}}, Vol.~\bibinfo{volume}{35}. \bibinfo{pages}{9630--9638}.
\newblock


\bibitem[\protect\citeauthoryear{Shin, Fowlkes, and Hoiem}{Shin et~al\mbox{.}}{2018}]%
        {shin2018pixels}
\bibfield{author}{\bibinfo{person}{Daeyun Shin}, \bibinfo{person}{Charless~C Fowlkes}, {and} \bibinfo{person}{Derek Hoiem}.} \bibinfo{year}{2018}\natexlab{}.
\newblock \showarticletitle{Pixels, voxels, and views: A study of shape representations for single view 3d object shape prediction}. In \bibinfo{booktitle}{\emph{Proceedings of the IEEE conference on computer vision and pattern recognition}}. \bibinfo{pages}{3061--3069}.
\newblock


\bibitem[\protect\citeauthoryear{Shin, Lee, Kim, and Kim}{Shin et~al\mbox{.}}{2017}]%
        {shin2017continual}
\bibfield{author}{\bibinfo{person}{Hanul Shin}, \bibinfo{person}{Jung~Kwon Lee}, \bibinfo{person}{Jaehong Kim}, {and} \bibinfo{person}{Jiwon Kim}.} \bibinfo{year}{2017}\natexlab{}.
\newblock \showarticletitle{Continual learning with deep generative replay}.
\newblock \bibinfo{journal}{\emph{Advances in neural information processing systems}}  \bibinfo{volume}{30} (\bibinfo{year}{2017}).
\newblock


\bibitem[\protect\citeauthoryear{Shmelkov, Schmid, and Alahari}{Shmelkov et~al\mbox{.}}{2017}]%
        {shmelkov2017incremental}
\bibfield{author}{\bibinfo{person}{Konstantin Shmelkov}, \bibinfo{person}{Cordelia Schmid}, {and} \bibinfo{person}{Karteek Alahari}.} \bibinfo{year}{2017}\natexlab{}.
\newblock \showarticletitle{Incremental learning of object detectors without catastrophic forgetting}. In \bibinfo{booktitle}{\emph{Proceedings of the IEEE international conference on computer vision}}. \bibinfo{pages}{3400--3409}.
\newblock


\bibitem[\protect\citeauthoryear{Smilkov, Thorat, Kim, Vi{\'e}gas, and Wattenberg}{Smilkov et~al\mbox{.}}{2017}]%
        {smilkov2017smoothgrad}
\bibfield{author}{\bibinfo{person}{Daniel Smilkov}, \bibinfo{person}{Nikhil Thorat}, \bibinfo{person}{Been Kim}, \bibinfo{person}{Fernanda Vi{\'e}gas}, {and} \bibinfo{person}{Martin Wattenberg}.} \bibinfo{year}{2017}\natexlab{}.
\newblock \showarticletitle{Smoothgrad: removing noise by adding noise}.
\newblock \bibinfo{journal}{\emph{arXiv preprint arXiv:1706.03825}} (\bibinfo{year}{2017}).
\newblock


\bibitem[\protect\citeauthoryear{Song and Xiao}{Song and Xiao}{2016}]%
        {song2016deep}
\bibfield{author}{\bibinfo{person}{Shuran Song} {and} \bibinfo{person}{Jianxiong Xiao}.} \bibinfo{year}{2016}\natexlab{}.
\newblock \showarticletitle{Deep sliding shapes for amodal 3d object detection in rgb-d images}. In \bibinfo{booktitle}{\emph{Proceedings of the IEEE conference on computer vision and pattern recognition}}. \bibinfo{pages}{808--816}.
\newblock


\bibitem[\protect\citeauthoryear{Srinivas and Fleuret}{Srinivas and Fleuret}{2019}]%
        {srinivas2019full}
\bibfield{author}{\bibinfo{person}{Suraj Srinivas} {and} \bibinfo{person}{Fran{\c{c}}ois Fleuret}.} \bibinfo{year}{2019}\natexlab{}.
\newblock \showarticletitle{Full-gradient representation for neural network visualization}.
\newblock \bibinfo{journal}{\emph{Advances in neural information processing systems}}  \bibinfo{volume}{32} (\bibinfo{year}{2019}).
\newblock


\bibitem[\protect\citeauthoryear{Tao, Hong, Chang, Dong, Wei, and Gong}{Tao et~al\mbox{.}}{2020}]%
        {tao2020few}
\bibfield{author}{\bibinfo{person}{Xiaoyu Tao}, \bibinfo{person}{Xiaopeng Hong}, \bibinfo{person}{Xinyuan Chang}, \bibinfo{person}{Songlin Dong}, \bibinfo{person}{Xing Wei}, {and} \bibinfo{person}{Yihong Gong}.} \bibinfo{year}{2020}\natexlab{}.
\newblock \showarticletitle{Few-shot class-incremental learning}. In \bibinfo{booktitle}{\emph{Proceedings of the IEEE/CVF Conference on Computer Vision and Pattern Recognition}}. \bibinfo{pages}{12183--12192}.
\newblock


\bibitem[\protect\citeauthoryear{Thai, Stojanov, Upadhya, and Rehg}{Thai et~al\mbox{.}}{2021}]%
        {thai20213d}
\bibfield{author}{\bibinfo{person}{Anh Thai}, \bibinfo{person}{Stefan Stojanov}, \bibinfo{person}{Vijay Upadhya}, {and} \bibinfo{person}{James~M Rehg}.} \bibinfo{year}{2021}\natexlab{}.
\newblock \showarticletitle{3d reconstruction of novel object shapes from single images}. In \bibinfo{booktitle}{\emph{2021 International Conference on 3D Vision (3DV)}}. IEEE, \bibinfo{pages}{85--95}.
\newblock


\bibitem[\protect\citeauthoryear{Van~der Maaten and Hinton}{Van~der Maaten and Hinton}{2008}]%
        {van2008visualizing}
\bibfield{author}{\bibinfo{person}{Laurens Van~der Maaten} {and} \bibinfo{person}{Geoffrey Hinton}.} \bibinfo{year}{2008}\natexlab{}.
\newblock \showarticletitle{Visualizing data using t-SNE.}
\newblock \bibinfo{journal}{\emph{Journal of machine learning research}} \bibinfo{volume}{9}, \bibinfo{number}{11} (\bibinfo{year}{2008}).
\newblock


\bibitem[\protect\citeauthoryear{Wallace and Hariharan}{Wallace and Hariharan}{2019}]%
        {wallace2019few}
\bibfield{author}{\bibinfo{person}{Bram Wallace} {and} \bibinfo{person}{Bharath Hariharan}.} \bibinfo{year}{2019}\natexlab{}.
\newblock \showarticletitle{Few-shot generalization for single-image 3d reconstruction via priors}. In \bibinfo{booktitle}{\emph{Proceedings of the IEEE/CVF international conference on computer vision}}. \bibinfo{pages}{3818--3827}.
\newblock


\bibitem[\protect\citeauthoryear{Wang and Fang}{Wang and Fang}{2020}]%
        {wang2020gsir}
\bibfield{author}{\bibinfo{person}{Jianren Wang} {and} \bibinfo{person}{Zhaoyuan Fang}.} \bibinfo{year}{2020}\natexlab{}.
\newblock \showarticletitle{Gsir: Generalizable 3d shape interpretation and reconstruction}. In \bibinfo{booktitle}{\emph{Computer Vision--ECCV 2020: 16th European Conference, Glasgow, UK, August 23--28, 2020, Proceedings, Part XIII 16}}. Springer, \bibinfo{pages}{498--514}.
\newblock


\bibitem[\protect\citeauthoryear{Wang, Zhang, Li, Fu, Liu, and Jiang}{Wang et~al\mbox{.}}{2018}]%
        {wang2018pixel2mesh}
\bibfield{author}{\bibinfo{person}{Nanyang Wang}, \bibinfo{person}{Yinda Zhang}, \bibinfo{person}{Zhuwen Li}, \bibinfo{person}{Yanwei Fu}, \bibinfo{person}{Wei Liu}, {and} \bibinfo{person}{Yu-Gang Jiang}.} \bibinfo{year}{2018}\natexlab{}.
\newblock \showarticletitle{Pixel2mesh: Generating 3d mesh models from single rgb images}. In \bibinfo{booktitle}{\emph{Proceedings of the European conference on computer vision (ECCV)}}. \bibinfo{pages}{52--67}.
\newblock


\bibitem[\protect\citeauthoryear{Wu, Wang, Xue, Sun, Freeman, and Tenenbaum}{Wu et~al\mbox{.}}{2017}]%
        {wu2017marrnet}
\bibfield{author}{\bibinfo{person}{Jiajun Wu}, \bibinfo{person}{Yifan Wang}, \bibinfo{person}{Tianfan Xue}, \bibinfo{person}{Xingyuan Sun}, \bibinfo{person}{Bill Freeman}, {and} \bibinfo{person}{Josh Tenenbaum}.} \bibinfo{year}{2017}\natexlab{}.
\newblock \showarticletitle{Marrnet: 3d shape reconstruction via 2.5 d sketches}.
\newblock \bibinfo{journal}{\emph{Advances in neural information processing systems}}  \bibinfo{volume}{30} (\bibinfo{year}{2017}).
\newblock


\bibitem[\protect\citeauthoryear{Wu, Zhang, Xue, Freeman, and Tenenbaum}{Wu et~al\mbox{.}}{2016}]%
        {wu2016learning}
\bibfield{author}{\bibinfo{person}{Jiajun Wu}, \bibinfo{person}{Chengkai Zhang}, \bibinfo{person}{Tianfan Xue}, \bibinfo{person}{Bill Freeman}, {and} \bibinfo{person}{Josh Tenenbaum}.} \bibinfo{year}{2016}\natexlab{}.
\newblock \showarticletitle{Learning a probabilistic latent space of object shapes via 3d generative-adversarial modeling}.
\newblock \bibinfo{journal}{\emph{Advances in neural information processing systems}}  \bibinfo{volume}{29} (\bibinfo{year}{2016}).
\newblock


\bibitem[\protect\citeauthoryear{Wu, Song, Khosla, Yu, Zhang, Tang, and Xiao}{Wu et~al\mbox{.}}{2015}]%
        {wu20153d}
\bibfield{author}{\bibinfo{person}{Zhirong Wu}, \bibinfo{person}{Shuran Song}, \bibinfo{person}{Aditya Khosla}, \bibinfo{person}{Fisher Yu}, \bibinfo{person}{Linguang Zhang}, \bibinfo{person}{Xiaoou Tang}, {and} \bibinfo{person}{Jianxiong Xiao}.} \bibinfo{year}{2015}\natexlab{}.
\newblock \showarticletitle{3d shapenets: A deep representation for volumetric shapes}. In \bibinfo{booktitle}{\emph{Proceedings of the IEEE conference on computer vision and pattern recognition}}. \bibinfo{pages}{1912--1920}.
\newblock


\bibitem[\protect\citeauthoryear{Xiang, Fu, Ji, and Huang}{Xiang et~al\mbox{.}}{2019}]%
        {xiang2019incremental}
\bibfield{author}{\bibinfo{person}{Ye Xiang}, \bibinfo{person}{Ying Fu}, \bibinfo{person}{Pan Ji}, {and} \bibinfo{person}{Hua Huang}.} \bibinfo{year}{2019}\natexlab{}.
\newblock \showarticletitle{Incremental learning using conditional adversarial networks}. In \bibinfo{booktitle}{\emph{Proceedings of the IEEE/CVF International Conference on Computer Vision}}. \bibinfo{pages}{6619--6628}.
\newblock


\bibitem[\protect\citeauthoryear{Xie, Yao, Sun, Zhou, and Zhang}{Xie et~al\mbox{.}}{2019}]%
        {xie2019pix2vox}
\bibfield{author}{\bibinfo{person}{Haozhe Xie}, \bibinfo{person}{Hongxun Yao}, \bibinfo{person}{Xiaoshuai Sun}, \bibinfo{person}{Shangchen Zhou}, {and} \bibinfo{person}{Shengping Zhang}.} \bibinfo{year}{2019}\natexlab{}.
\newblock \showarticletitle{Pix2vox: Context-aware 3d reconstruction from single and multi-view images}. In \bibinfo{booktitle}{\emph{Proceedings of the IEEE/CVF international conference on computer vision}}. \bibinfo{pages}{2690--2698}.
\newblock


\bibitem[\protect\citeauthoryear{Xing, Chen, Ling, Zhou, and Xiang}{Xing et~al\mbox{.}}{2022a}]%
        {xing2022few}
\bibfield{author}{\bibinfo{person}{Zhen Xing}, \bibinfo{person}{Yijiang Chen}, \bibinfo{person}{Zhixin Ling}, \bibinfo{person}{Xiangdong Zhou}, {and} \bibinfo{person}{Yu Xiang}.} \bibinfo{year}{2022}\natexlab{a}.
\newblock \showarticletitle{Few-shot single-view 3d reconstruction with memory prior contrastive network}. In \bibinfo{booktitle}{\emph{European Conference on Computer Vision}}. Springer, \bibinfo{pages}{55--70}.
\newblock


\bibitem[\protect\citeauthoryear{Xing, Li, Wu, and Jiang}{Xing et~al\mbox{.}}{2022b}]%
        {xing2022semi}
\bibfield{author}{\bibinfo{person}{Zhen Xing}, \bibinfo{person}{Hengduo Li}, \bibinfo{person}{Zuxuan Wu}, {and} \bibinfo{person}{Yu-Gang Jiang}.} \bibinfo{year}{2022}\natexlab{b}.
\newblock \showarticletitle{Semi-supervised single-view 3d reconstruction via prototype shape priors}. In \bibinfo{booktitle}{\emph{European Conference on Computer Vision}}. Springer, \bibinfo{pages}{535--551}.
\newblock


\bibitem[\protect\citeauthoryear{Xu and Zhu}{Xu and Zhu}{2018}]%
        {xu2018reinforced}
\bibfield{author}{\bibinfo{person}{Ju Xu} {and} \bibinfo{person}{Zhanxing Zhu}.} \bibinfo{year}{2018}\natexlab{}.
\newblock \showarticletitle{Reinforced continual learning}.
\newblock \bibinfo{journal}{\emph{Advances in Neural Information Processing Systems}}  \bibinfo{volume}{31} (\bibinfo{year}{2018}).
\newblock


\bibitem[\protect\citeauthoryear{Yan, Tian, Shi, Guo, Wang, and Zha}{Yan et~al\mbox{.}}{2021}]%
        {yan2021continual}
\bibfield{author}{\bibinfo{person}{Zike Yan}, \bibinfo{person}{Yuxin Tian}, \bibinfo{person}{Xuesong Shi}, \bibinfo{person}{Ping Guo}, \bibinfo{person}{Peng Wang}, {and} \bibinfo{person}{Hongbin Zha}.} \bibinfo{year}{2021}\natexlab{}.
\newblock \showarticletitle{Continual neural mapping: Learning an implicit scene representation from sequential observations}. In \bibinfo{booktitle}{\emph{Proceedings of the IEEE/CVF International Conference on Computer Vision}}. \bibinfo{pages}{15782--15792}.
\newblock


\bibitem[\protect\citeauthoryear{Yang, Cui, Belongie, and Hariharan}{Yang et~al\mbox{.}}{2018}]%
        {yang2018learning}
\bibfield{author}{\bibinfo{person}{Guandao Yang}, \bibinfo{person}{Yin Cui}, \bibinfo{person}{Serge Belongie}, {and} \bibinfo{person}{Bharath Hariharan}.} \bibinfo{year}{2018}\natexlab{}.
\newblock \showarticletitle{Learning single-view 3d reconstruction with limited pose supervision}. In \bibinfo{booktitle}{\emph{Proceedings of the European Conference on Computer Vision (ECCV)}}. \bibinfo{pages}{86--101}.
\newblock


\bibitem[\protect\citeauthoryear{Yoon, Yang, Lee, and Hwang}{Yoon et~al\mbox{.}}{2017}]%
        {yoon2017lifelong}
\bibfield{author}{\bibinfo{person}{Jaehong Yoon}, \bibinfo{person}{Eunho Yang}, \bibinfo{person}{Jeongtae Lee}, {and} \bibinfo{person}{Sung~Ju Hwang}.} \bibinfo{year}{2017}\natexlab{}.
\newblock \showarticletitle{Lifelong learning with dynamically expandable networks}.
\newblock \bibinfo{journal}{\emph{arXiv preprint arXiv:1708.01547}} (\bibinfo{year}{2017}).
\newblock


\bibitem[\protect\citeauthoryear{Yun, Liu, and Liu}{Yun et~al\mbox{.}}{2021}]%
        {yun2021defense}
\bibfield{author}{\bibinfo{person}{Peng Yun}, \bibinfo{person}{Yuxuan Liu}, {and} \bibinfo{person}{Ming Liu}.} \bibinfo{year}{2021}\natexlab{}.
\newblock \showarticletitle{In defense of knowledge distillation for task incremental learning and its application in 3d object detection}.
\newblock \bibinfo{journal}{\emph{IEEE Robotics and Automation Letters}} \bibinfo{volume}{6}, \bibinfo{number}{2} (\bibinfo{year}{2021}), \bibinfo{pages}{2012--2019}.
\newblock


\bibitem[\protect\citeauthoryear{Zenke, Poole, and Ganguli}{Zenke et~al\mbox{.}}{2017}]%
        {zenke2017continual}
\bibfield{author}{\bibinfo{person}{Friedemann Zenke}, \bibinfo{person}{Ben Poole}, {and} \bibinfo{person}{Surya Ganguli}.} \bibinfo{year}{2017}\natexlab{}.
\newblock \showarticletitle{Continual learning through synaptic intelligence}. In \bibinfo{booktitle}{\emph{International conference on machine learning}}. PMLR, \bibinfo{pages}{3987--3995}.
\newblock


\bibitem[\protect\citeauthoryear{Zhang, B{\"u}tepage, Kjellstr{\"o}m, and Mandt}{Zhang et~al\mbox{.}}{2018a}]%
        {zhang2018advances}
\bibfield{author}{\bibinfo{person}{Cheng Zhang}, \bibinfo{person}{Judith B{\"u}tepage}, \bibinfo{person}{Hedvig Kjellstr{\"o}m}, {and} \bibinfo{person}{Stephan Mandt}.} \bibinfo{year}{2018}\natexlab{a}.
\newblock \showarticletitle{Advances in variational inference}.
\newblock \bibinfo{journal}{\emph{IEEE transactions on pattern analysis and machine intelligence}} \bibinfo{volume}{41}, \bibinfo{number}{8} (\bibinfo{year}{2018}), \bibinfo{pages}{2008--2026}.
\newblock


\bibitem[\protect\citeauthoryear{Zhang, Song, Lin, Zheng, Pan, and Xu}{Zhang et~al\mbox{.}}{2021}]%
        {zhang2021few}
\bibfield{author}{\bibinfo{person}{Chi Zhang}, \bibinfo{person}{Nan Song}, \bibinfo{person}{Guosheng Lin}, \bibinfo{person}{Yun Zheng}, \bibinfo{person}{Pan Pan}, {and} \bibinfo{person}{Yinghui Xu}.} \bibinfo{year}{2021}\natexlab{}.
\newblock \showarticletitle{Few-shot incremental learning with continually evolved classifiers}. In \bibinfo{booktitle}{\emph{Proceedings of the IEEE/CVF conference on computer vision and pattern recognition}}. \bibinfo{pages}{12455--12464}.
\newblock


\bibitem[\protect\citeauthoryear{Zhang, Candra, Vetter, and Zakhor}{Zhang et~al\mbox{.}}{2015}]%
        {zhang2015sensor}
\bibfield{author}{\bibinfo{person}{Richard Zhang}, \bibinfo{person}{Stefan~A Candra}, \bibinfo{person}{Kai Vetter}, {and} \bibinfo{person}{Avideh Zakhor}.} \bibinfo{year}{2015}\natexlab{}.
\newblock \showarticletitle{Sensor fusion for semantic segmentation of urban scenes}. In \bibinfo{booktitle}{\emph{2015 IEEE international conference on robotics and automation (ICRA)}}. IEEE, \bibinfo{pages}{1850--1857}.
\newblock


\bibitem[\protect\citeauthoryear{Zhang and Rudnicky}{Zhang and Rudnicky}{2006}]%
        {zhang2006new}
\bibfield{author}{\bibinfo{person}{Rong Zhang} {and} \bibinfo{person}{Alexander~I Rudnicky}.} \bibinfo{year}{2006}\natexlab{}.
\newblock \showarticletitle{A new data selection principle for semi-supervised incremental learning}. In \bibinfo{booktitle}{\emph{18th International Conference on Pattern Recognition (ICPR'06)}}, Vol.~\bibinfo{volume}{2}. IEEE, \bibinfo{pages}{780--783}.
\newblock


\bibitem[\protect\citeauthoryear{Zhang, Zhang, Zhang, Tenenbaum, Freeman, and Wu}{Zhang et~al\mbox{.}}{2018b}]%
        {zhang2018learning}
\bibfield{author}{\bibinfo{person}{Xiuming Zhang}, \bibinfo{person}{Zhoutong Zhang}, \bibinfo{person}{Chengkai Zhang}, \bibinfo{person}{Josh Tenenbaum}, \bibinfo{person}{Bill Freeman}, {and} \bibinfo{person}{Jiajun Wu}.} \bibinfo{year}{2018}\natexlab{b}.
\newblock \showarticletitle{Learning to reconstruct shapes from unseen classes}.
\newblock \bibinfo{journal}{\emph{Advances in neural information processing systems}}  \bibinfo{volume}{31} (\bibinfo{year}{2018}).
\newblock


\bibitem[\protect\citeauthoryear{Zhou, Khosla, Lapedriza, Oliva, and Torralba}{Zhou et~al\mbox{.}}{2016}]%
        {zhou2016learning}
\bibfield{author}{\bibinfo{person}{Bolei Zhou}, \bibinfo{person}{Aditya Khosla}, \bibinfo{person}{Agata Lapedriza}, \bibinfo{person}{Aude Oliva}, {and} \bibinfo{person}{Antonio Torralba}.} \bibinfo{year}{2016}\natexlab{}.
\newblock \showarticletitle{Learning deep features for discriminative localization}. In \bibinfo{booktitle}{\emph{Proceedings of the IEEE conference on computer vision and pattern recognition}}. \bibinfo{pages}{2921--2929}.
\newblock


\end{thebibliography}

\appendix

\clearpage

\section{Algorithm:}
Our training process is divided in 3 parts training testing and inference.
First we describe the incremental training of the shapes
\begin{algorithm}[H]
\caption{\textsc{IncrementalSessionLoss}: Loss update at each incremental training session}
\textbf{Input:}Training set $\mathcal{D}^{i}$=$\{(\mathcal{I}_{1},\mathcal{S}_{1}),...,(\mathcal{I}_{|\mathcal{D}^{i}|},\mathcal{S}_{|\mathcal{D}^{i}|})\}$ and $\mathcal{M}_{\mathrm{SAL}}$.

\textbf{Output:} The loss $\mathcal{L}$ for a randomly generated training session
\begin{algorithmic}
\State $\mathcal{L} \gets 0$ \Comment{Initialize loss}

\For{$k \in \{1,...,\mathcal{N}^C\}$} 
        \State $\mathcal{P} \gets \mathcal{M}_{\mathrm{SAL}}$ \Comment{Load from saliency memory bank}
        \State $\mathcal{D} \gets \mathcal{D}^i \bigoplus \mathcal{P}$
        
        \State $ \mathcal{L} \gets \textsc{BinaryCrossEntropy}(\mathcal{D})$
        \State $\mathcal{L} \gets \mathcal{L} + \mathrm{KL}(\mathcal{Q}(t)||\Tilde{\mathcal{Q}}(t-1))$
\EndFor
\end{algorithmic}
\label{alg:ALG2}
\end{algorithm}

We test the model performance on the cumulative dataset $\mathcal{C} = \{\mathcal{C}_1, \mathcal{C}_2, \mathcal{C}_i\}$

Code is available at \href{https://drive.google.com/file/d/15o_rUGAkEwXw8Yw-39t8aQZ5Hf1GJJ56/view?usp=sharing}{this link}.

\section{Saliency map:} 
We utilize attention saliency maps to generate binary maps that emphasize the regions of interest within each image. To derive these attention maps, we introduce additional layers into the 2D image encoder, applying a procedure as described below.
Starting with the feature map extractor's output, we aggregate it into a vector represented as $\mathcal{L}^s = {l_1^{s}, l_2^{s}, \dots l_n^{s}}$. The global feature vector $\mathcal{G}$ is generated by applying nonlinearity to the output feature vector, producing the output class probability. Using this, we use a compatibility function $\mathcal{C}$ that calculates a compatibility score between two vectors of the same dimensions. Thus, we obtain compatibility functions $\mathcal{C}(\Tilde{\mathcal{L}}^s, \mathcal{G})={c_1^s, c_2^s, \dots c_n^s}$, where $\Tilde{\mathcal{L}}^s$ corresponds to $\mathcal{L}^s$ with the dimension of $\mathcal{G}$. To normalize the compatibility scores, we employ the \texttt{softmax} function, resulting in the normalized attention vector $\mathcal{A}^s = {\mathcal{A}_1^s, \mathcal{A}_2^s, \dots \mathcal{A}_n^s}$. Subsequently, we generate the global vector $\mathcal{G}$ from the normalized attention vectors using the following equation:
$ \mathcal{G}^s = \sum_i \mathcal{A}_i^s \cdot l_i^s$.
These newly obtained global descriptor vectors are concatenated into a vector and utilized in a classification layer to calculate the output probabilities. The attention layer's parameters are trained simultaneously with the feature extractor. Alternatively, if feasible, the attention layer can be trained separately along with an additional feature extractor.

\begin{figure}
  \centering
   \includegraphics[width=1.0\linewidth]{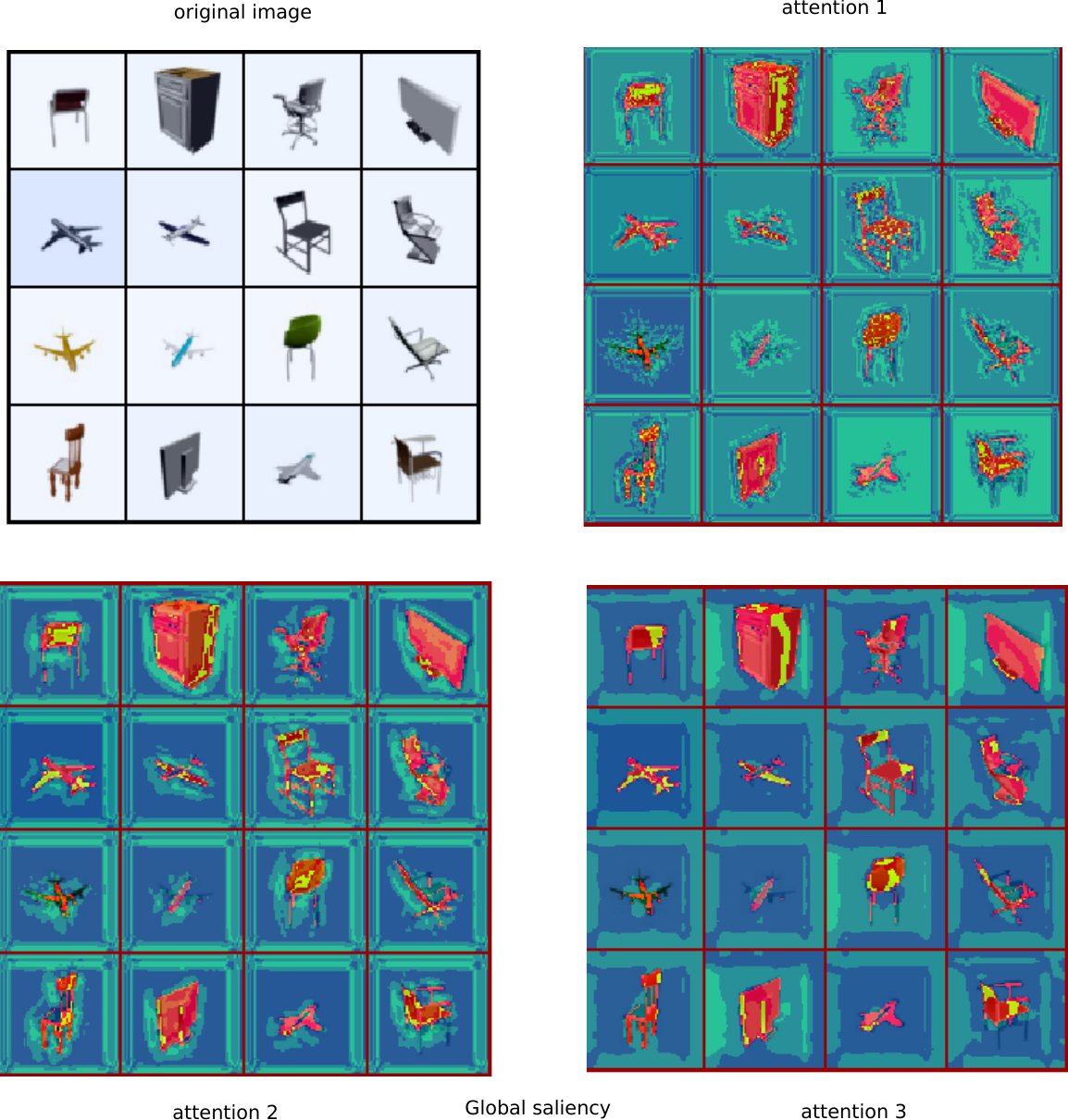}
   \caption{\footnotesize{Saliency map of different objects at three different layers.}}
   \label{fig:saliency_map_supp_1}
\end{figure}

\begin{figure}
  \centering
   \includegraphics[width=1.0\linewidth]{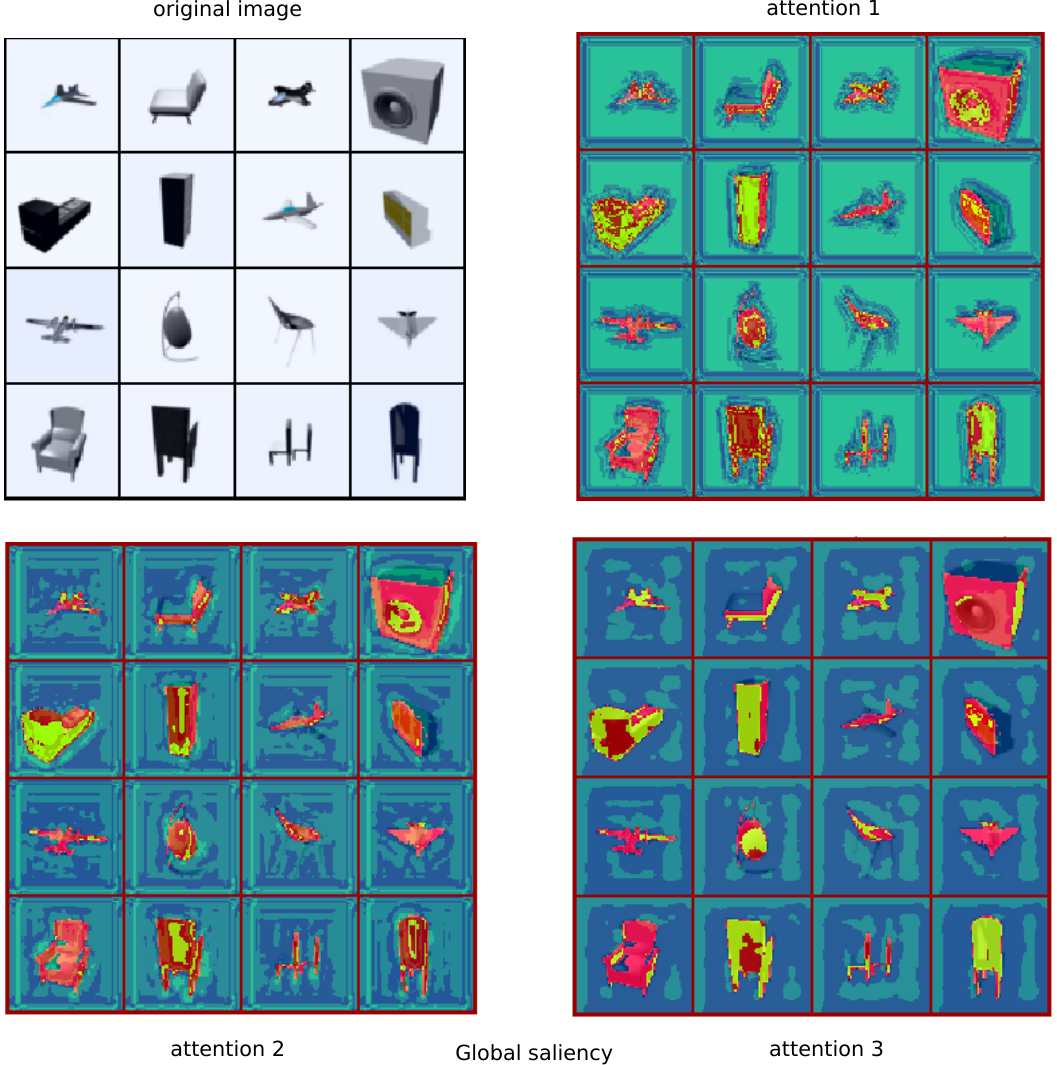}
   \caption{\footnotesize{Saliency map of different objects at three different layers.}}
   \label{fig:saliency_map_supp_2}
\end{figure}

\begin{figure}
  \centering
   \includegraphics[width=1.0\linewidth]{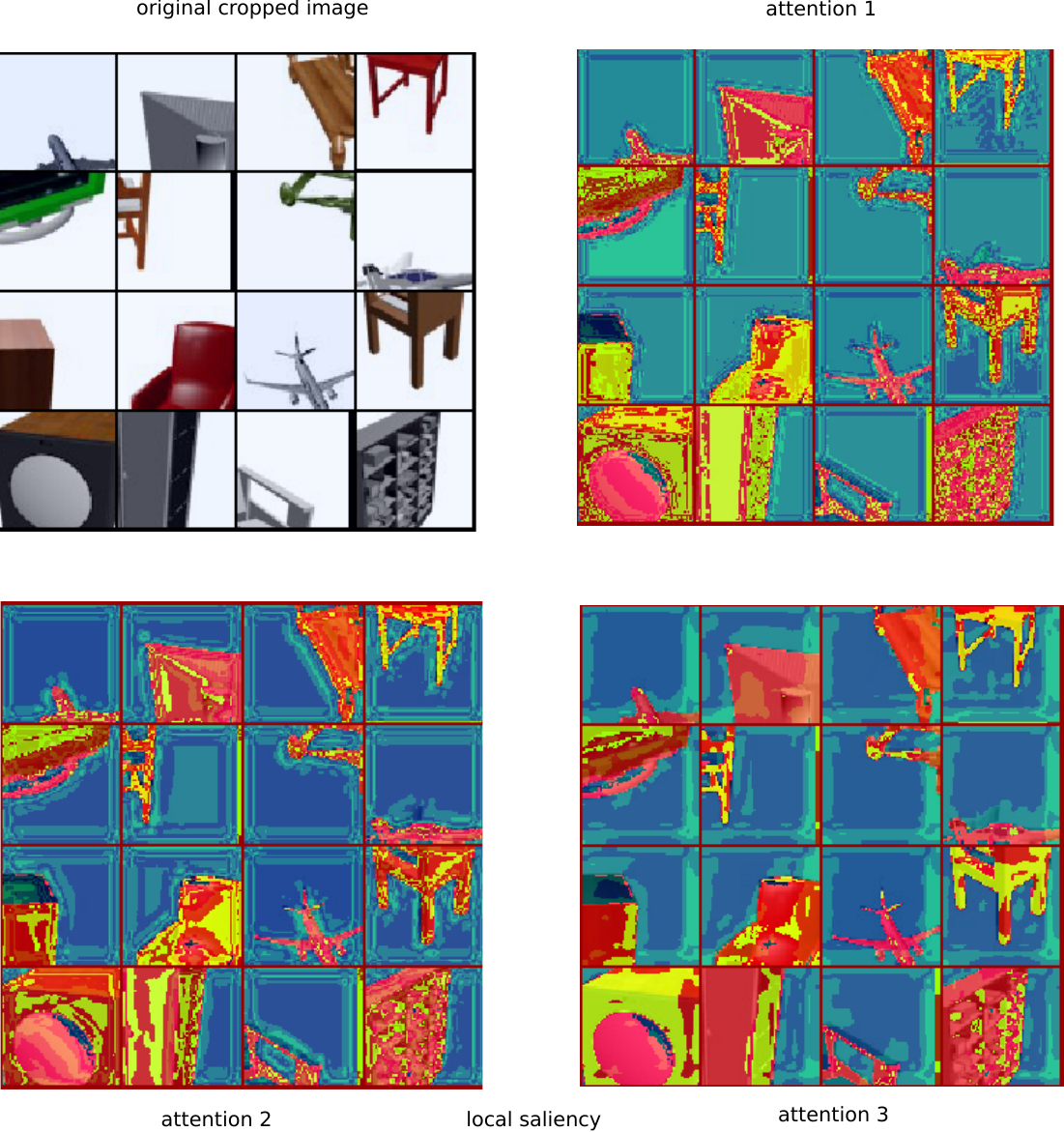}
   \caption{\footnotesize{Local Saliency map of different objects at three different layers.}}
   \label{fig:local_saliency_map}
\end{figure}

The compatibility function can assume either a dot-product or additive form, contingent on its interaction with the global feature vector $\mathcal{G}$:
$ \mathcal{C} = <\mathcal{L}, \mathcal{G}> =\mathcal{L} + \mathcal{G}$
or,
$\mathcal{C} = <\mathcal{L}, \mathcal{G}> = \mathcal{L} \times \mathcal{G}$.
Both of these fall under the category of pairwise attention. The parameters of the attention layer are learned through cross-entropy loss. We employ three layers of spatial attention. The resulting attention for global saliency for these three layers is visually demonstrated in Fig.\ref{fig:saliency_map_supp_1} and Fig.\ref{fig:saliency_map_supp_2}. The local saliency map is shown in Fig.~\ref{fig:local_saliency_map}.
We retain the corresponding attention saliency maps along with their corresponding ground truth for each session. These images play a crucial role during incremental training, aiding the model in retaining knowledge from previous sessions.

\begin{table}
        \centering
        \begin{tabular}{c c c c c c c}
        \hline
        \textbf{Object}&\multicolumn{5}{c}{\textbf{Session}}\\
        \cline{2-6} 
        \textbf{} & \textbf{0}& \textbf{1}& \textbf{2}& \textbf{3}& \textbf{4} & \textbf{mean}$\uparrow$ \\
        \hline
        Loudspeaker & \textbf{0.662}
         & 0.621 & 0.615 & 0.604 & 0.586 \\
        Display & \textbf{0.471}
         & 0.456 & 0.441 & 0.430 & 0.426 \\
        Airplane & \textbf{0.571} & 0.546 & 0.532 & 0.521 & 0.517 & \textbf{0.532}\\
        Cabinet & \textbf{0.733}
         & 0.709 & 0.702 & 0.698 & 0.681 \\
        Chair & \textbf{0.501}
         & 0.475 & 0.464 & 0.456 & 0.448 \\
        \hline
        Rifle & - 
         & \textbf{0.486} & 0.452 & 0.446 & 0.431 \\
        Table & -
         & \textbf{0.506} & 0.487 & 0.479 & 0.466 & \textbf{0.449}\\
        \hline
        Car & -
         & - & \textbf{0.742} & 0.715 & 0.709 \\
        Sofa 
         & -
         & - & \textbf{0.681} & 0.667 & 0.653 & \textbf{0.681}\\
        \hline
        Bench & -
         & - & - & \textbf{0.485} & 0.465 \\
        Lamp & -
         & - & - & \textbf{0.371} & 0.352 & \textbf{0.409}\\
        \hline
        Telephone 
        & -
         & - & - & - & \textbf{0.719} \\
        Vessel & -
         & - & - & - & \textbf{0.527} & \textbf{0.623}\\
        \hline
        \end{tabular}
        \caption{Variation in Intersection over Union (IOU) values for distinct objects in the ShapeNet dataset across multiple sessions for the combination of occupancy network and EWC.}
        \label{table:IOU_var_shapenet_ewc}
        \end{table}
        
        \begin{table}
        \centering
        \begin{tabular}{c c c c c c c}
        \hline
        \textbf{Object}&\multicolumn{5}{c}{\textbf{Session}}\\
        \cline{2-6} 
        \textbf{} & \textbf{0}& \textbf{1}& \textbf{2}& \textbf{3}& \textbf{4}& \textbf{mean}$\uparrow$ \\
        \hline
        Loudspeaker & \textbf{0.662}
         & 0.615 & 0.609 & 0.597 & 0.587\\
        Display & \textbf{0.471}
         & 0.446 & 0.433 & 0.421 & 0.411\\
        Airplane & \textbf{0.571} & 0.532 & 0.520 & 0.514 & 0.506 &\textbf{0.521}\\
        Cabinet & \textbf{0.733}
         & 0.702 & 0.691 & 0.684 & 0.671\\
         Chair &\textbf{0.501}
         & 0.474 & 0.467 & 0.448& 0.437\\
        \hline
        Rifle & - 
         & \textbf{0.486} & 0.448 & 0.434& 0.425 \\
        Table & - 
         & \textbf{0.506} & 0.472 & 0.463& 0.454& \textbf{0.44}\\
        \hline
        Car & -
         & - & \textbf{0.742} & 0.703 & 0.692\\
        Sofa & -
         & - & \textbf{0.681} & 0.654& 0.642& \textbf{0.667}\\
        \hline
        Bench 
         & -
         & - & - & \textbf{0.485}& 0.453\\
        Lamp & -
         & - & - & \textbf{0.371}& 0.345& \textbf{0.399}  \\
        \hline
        Telephone 
         & -
         & - & - & - &  \textbf{0.719}\\
        Vessel & -
         & - & - & - & \textbf{0.527}& \textbf{0.623}  \\
        \hline
        \end{tabular}
        \caption{Variation in Intersection over Union (IOU) values for distinct objects in the ShapeNet dataset across multiple sessions for the combination of occupancy network and GEM.}
        \label{table:IOU_var_shapenet_gem}
        \end{table}


\subsection{Local saliency insertion:}

In this section, we elucidate the various processes that have been integrated into Section~\ref{sec:ablation}. In the context of the image regeneration process, we have explored three distinct strategies to address how cropped patches are arranged. These strategies are essential for maintaining the spatial and saliency information of the original image during the regeneration process. Let's delve into each of these strategies in more detail:

\textbf{Placing All Patches with Coordinate Preservation:} The first strategy involves placing all the cropped patches back in their original positions within the image. This means that the coordinates of the patches are maintained as they were in the original image. By doing so, the spatial relationships between different parts of the image are preserved. This strategy is crucial for ensuring that the regenerated image maintains its structure and appearance.

\textbf{Zero-Padding for Desired Shape:} In this strategy, the goal is to achieve a specific shape for the regenerated image. To achieve this, the cropped patches are zero-padded before being reconstructed into the image. This is particularly useful when the patches don't perfectly fit the desired shape, and zero-padding helps in filling the gaps to achieve the intended dimensions. Zero-padding ensures that the regenerated image conforms to a consistent and predefined size.

\textbf{Compression and Compressed Saliency Map Storage:} This strategy involves compressing the original image and storing the compressed saliency map, which captures the essential information about regions of interest and their significance. During rehearsal, when regenerating the image, the compressed saliency map is used to guide the interpolation process, ensuring that important features are retained.

    \begin{table}
        \centering
        \begin{tabular}{c c c c c c c}
        \hline
        \textbf{Object}&\multicolumn{5}{c}{\textbf{Session}}\\
        \cline{2-6} 
        \textbf{} & \textbf{0}& \textbf{1}& \textbf{2}& \textbf{3}& \textbf{4} & \textbf{mean}$\uparrow$ \\
        \hline
        Loudspeaker & \textbf{0.662}
         & 0.617 & 0.605 & 0.594 & 0.569 \\
        Display & \textbf{0.471}
         & 0.436 & 0.423 & 0.415 & 0.402 \\
        Airplane & \textbf{0.571} & 0.536 & 0.527 & 0.515 & 0.504 & \textbf{0.517}\\
        Cabinet & \textbf{0.733}
         & 0.715 & 0.703 & 0.692 & 0.669 \\
        Chair & \textbf{0.501}
         & 0.476 & 0.462 & 0.451 & 0.438 \\
        \hline
        Rifle & - 
         & \textbf{0.486} & 0.442 & 0.428 & 0.419 \\
        Table & -
         & \textbf{0.506} & 0.467 & 0.458 & 0.441 & \textbf{0.43}\\
        \hline
        Car & -
         & - & \textbf{0.742} & 0.711 & 0.702 \\
        Sofa 
         & -
         & - & \textbf{0.681} & 0.652 & 0.643 & \textbf{0.672}\\
        \hline
        Bench & -
         & - & - & \textbf{0.485} & 0.463 \\
        Lamp & -
         & - & - & \textbf{0.371} & 0.351 & \textbf{0.407}\\
        \hline
        Telephone 
        & -
         & - & - & - & \textbf{0.719} \\
        Vessel & -
         & - & - & - & \textbf{0.527} & \textbf{0.623}\\
        \hline
        \end{tabular}
        \caption{Variation in Intersection over Union (IOU) values for distinct objects in the ShapeNet dataset across multiple sessions for the combination of occupancy network and DEN.}
        \label{table:IOU_var_shapenet_den}
        \end{table}

\section{Generate mesh from latent module:}
In our approach, we employ a specialized module called Multiresolution Isosurface Extraction to generate a mesh from the latent module's output. The goal of this module is to create a 3D representation of the object's shape based on the information provided by the latent variables. The process begins with obtaining the output probabilities for each voxel. These probabilities indicate whether a particular voxel should be considered occupied by the object. To make this determination, a threshold is applied to the probability values. If a voxel's probability exceeds this threshold, it is marked as occupied, implying that the object is present at that location in space. Subsequently, we utilize a technique known as the marching cubes algorithm. This algorithm transforms the 3D voxel data, specifically the information about occupied voxels, into a smooth and continuous surface representation. The algorithm identifies the grid cells that intersect the object's surface and then creates polygons that approximate the shape of the object within those cells. These polygons collectively form the desired isosurface, which represents the outer boundary of the object in 3D space. 

Similarly, an alternative method that can be applied within a comparable context is the utilization of the Marching Tetrahedra algorithm. The Marching Tetrahedra algorithm extends the concept of the marching cubes algorithm by operating on tetrahedral cells instead of cubes. This makes it more suitable for irregularly sampled data and provides more accurate representations of complex shapes.  
In this method, each tetrahedron is formed by connecting the four corner points of a cell in the voxel grid. The algorithm evaluates the scalar field (in this case, the occupancy probabilities) at the vertices of each tetrahedron. Based on these scalar values, the algorithm determines whether a surface intersects a tetrahedron and generates triangles accordingly. The main advantage of the Marching Tetrahedra algorithm is its ability to handle higher-order interpolations and adapt to varying resolutions in the voxel data. This can result in more accurate and detailed mesh representations, especially for objects with intricate shapes and structures.

\section{Additional Results:}
In this section, we provide a comprehensive overview of the distinct experimental setups detailed in Section \ref{sec:experiments} and present the corresponding outcomes.

\subsection{Integrating EWC with Occupancy network:}
The process starts by identifying the network parameters that necessitate regularization through EWC. In the context of an occupancy network, these parameters encompass the weights and biases of the encoder and decoder layers. For each parameter, the Fisher information associated with the previous tasks is computed. The Fisher information gauges the parameter's contribution to the loss of a specific task and is calculated using the gradients of the loss with respect to that parameter. Given the nature of single-image 3D reconstruction tasks, the Fisher information is estimated based on data from the preceding tasks. When training on a new task, the EWC regularization term is incorporated into the loss function. This term for each parameter is directly proportional to the Fisher information and the squared difference between the current parameter value and its value after training on the previous task. The complete loss function combines the reconstruction loss for the current task and the EWC regularization terms for the previous tasks. The balance between these two components is modulated to control the trade-off between learning the new task and retaining knowledge from the earlier ones. Thus, for every new task, the Fisher information for the previous tasks is computed, the EWC regularization terms are added to the loss function, and the model's parameters are updated using gradient descent. The outcome of this integration is presented in Table \ref{table:IOU_var_shapenet_ewc}.

\subsection{Integrating DEN with occupancy network:}
 This fusion incorporates the principles of Dynamic Expandable Networks by introducing mechanisms that enable the network's architecture to adapt and adjust based on the complexity of tasks and the characteristics of the data. DEN establishes guidelines for determining when the network's architecture should expand or contract. Specifically, when confronted with new object classes or intricate shapes that pose a challenge, the network autonomously decides to expand its architecture to capture finer details and accommodate the new task's demands. 
 We extend the network's architecture by incorporating supplementary layers and units to accommodate the assimilation of new knowledge. Additionally, the integration also necessitates the implementation of mechanisms to efficiently manage computational resources. This entails the introduction of proper contraction-expansion strategies. In situations where tasks exhibit simplicity or mitigate the risk of overfitting or excessive resource utilization, the Dynamic Expandable Network (DEN) contracts its architecture by eliminating units or layers. This result is shown in Fig.~\ref{table:IOU_var_shapenet_den}.

\subsection{Integrating GEM with occupancy network:}
The integration of Gradient Episodic Memory (GEM) with an Occupancy Network for continual 3D reconstruction entails the incorporation of GEM to address the challenge of catastrophic forgetting while preserving the knowledge gained from prior tasks. GEM operates by updating gradients within a feasible region, delineated by samples stored in an episodic memory. This memory module retains a subset of data from earlier tasks and ensures that the gradients pertaining to the new task do not trigger substantial alterations in predictions for the retained data. The process involves memory mechanisms for the storage and management of data samples from previous tasks. GEM maintains a memory buffer with a predefined capacity to store input-output pairs. During each incremental session, after training on the new task, this replays a subset of stored data from the memory buffer to ensure that the model retains its performance on previous tasks. To implement this, adjustments are made to the loss function to incorporate GEM's regularization term, which imposes constraints on gradient updates pertaining to the stored data. Various parameters of GEM, such as the buffer size and regularization strength, are modified to optimize performance and memory utilization. The optimal values for these parameters are determined through experimentation to achieve the best possible results. The outcomes of integrating GEM with the Occupancy Network are presented in Table \ref{table:IOU_var_shapenet_gem}, showcasing the impact of this approach on the variability of Intersection over Union (IOU) scores across different epochs and tasks within the ShapeNet dataset.

\end{document}